\documentclass[10pt,twocolumn,letterpaper]{article}

\usepackage{cvpr}              
\usepackage{algorithm}
\usepackage{algpseudocode}

\usepackage[dvipsnames]{xcolor}

\definecolor{cvprblue}{rgb}{0.21,0.49,0.74}
\usepackage[pagebackref,breaklinks,colorlinks,citecolor=cvprblue]{hyperref}

\usepackage[utf8]{inputenc} 
\usepackage[T1]{fontenc}    
\usepackage{hyperref}       
\usepackage{url}            
\usepackage{booktabs}       
\usepackage{amsfonts}       
\usepackage{nicefrac}       
\usepackage{microtype}      
\usepackage{xcolor}         
\usepackage{tikz}

\usepackage[accsupp]{axessibility} 

\usepackage{multirow, array}
\usepackage{tabularx,tabulary, booktabs, colortbl}
\usepackage[overload]{ragged2e}

\usepackage{makecell}

\usepackage[most]{tcolorbox}  
\usepackage{subcaption}

\usepackage{xspace}
\usepackage{float}

\definecolor{lightblue}{rgb}{0.1,0.3,1}
\definecolor{lightgreen}{rgb}{0.1,0.8,0.1}
\definecolor{defaultcolor}{gray}{0.9}
\newlength\savewidth\newcommand\shline{\noalign{\global\savewidth\arrayrulewidth
  \global\arrayrulewidth 1pt}\hline\noalign{\global\arrayrulewidth\savewidth}}
\newcommand{\tablestyle}[2]{\setlength{\tabcolsep}{#1}\renewcommand{\arraystretch}{#2}\centering\footnotesize}

\newcommand{\eat}[1]{\ignorespaces}
\newcommand{\dataset}[0]{\textsc{Quilt}\xspace}
\newcommand{\quiltinstruct}{\textsc{Quilt-Instruct}\xspace}
\newcommand{\quiltvqa}{\textsc{Quilt-VQA}\xspace}
\newcommand{\model}[0]{\textsc{Quilt-LLaVA}\xspace}
\newcommand{\quiltmodel}[0]{\textsc{QuiltNet}\xspace}
\newcommand{\llava}[0]{\textsc{LLaVA}\xspace}
\newcommand{\llavamed}[0]{\textsc{LLaVA-MED}\xspace}
\newcommand{\halfinstruct}[0]{\textsc{40K}\xspace}
\newcommand{\fullinstruct}[0]{\textsc{107K}\xspace} 

\def\paperID{1897} 
\def\confName{CVPR}
\def\confYear{2024}

\title{\includegraphics[height=25pt]{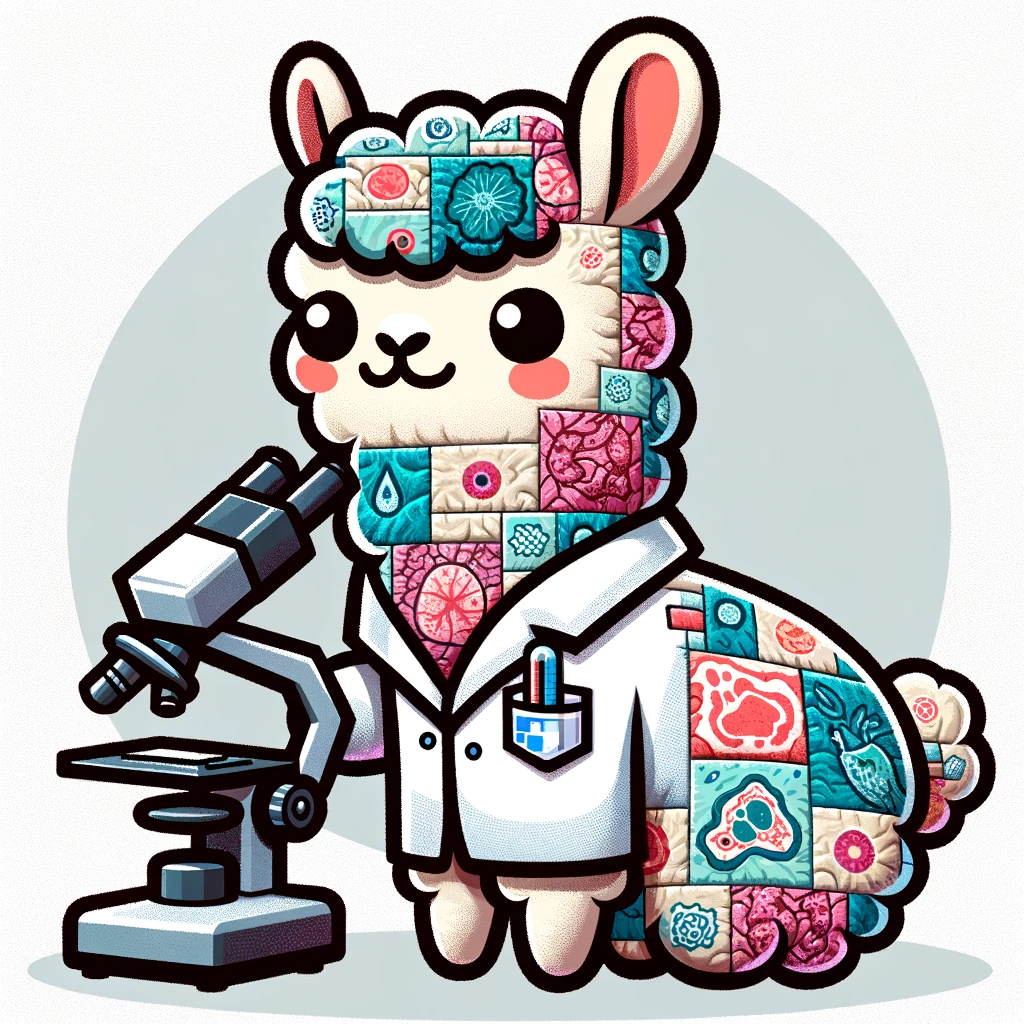}
Quilt-LLaVA: Visual Instruction Tuning by Extracting Localized Narratives from Open-Source Histopathology Videos
}

\author{
Mehmet Saygin Seyfioglu$^{\spadesuit, \ddagger}$\thanks{msaygin@cs.washington.edu; $^{\spadesuit}$ Denotes equal contribution.} \qquad Wisdom O. Ikezogwo$^{\spadesuit, \ddagger}$ \qquad Fatemeh Ghezloo$^{\spadesuit, \ddagger}$
\\
Ranjay Krishna$^{\dagger, \ddagger}$ \qquad Linda Shapiro$^{\ddagger}$
\\
\normalsize
$^{\ddagger}$University of Washington \\
\normalsize
$^{\dagger}$Allen Institute for AI (AI2)
}

\begin{document}
\maketitle
\begin{abstract}

Diagnosis in histopathology requires a global whole slide images (WSIs) analysis, requiring pathologists to compound evidence from different WSI patches.
The gigapixel scale of WSIs poses a challenge for histopathology multi-modal models. 
Training multi-model models for histopathology requires instruction tuning datasets, which currently contain information for individual image patches, without a spatial grounding of the concepts within each patch and without a wider view of the WSI.
To bridge this gap, we introduce \quiltinstruct, a large-scale dataset of $107,131$ histopathology-specific instruction question/answer pairs, grounded within diagnostically relevant image patches that make up the WSI. Our dataset is collected by leveraging educational histopathology videos from YouTube, which provides spatial localization of narrations by automatically extracting the narrators' cursor positions. 
\quiltinstruct supports contextual reasoning by extracting diagnosis and supporting facts from the entire WSI. 
Using \quiltinstruct, we train \model, which can reason beyond the given single image patch, enabling diagnostic reasoning across patches. To evaluate \model, we propose a comprehensive evaluation dataset created from $985$ images and $1283$ human-generated question-answers. We also thoroughly evaluate \model using public histopathology datasets, where \model significantly outperforms SOTA by over 10\% on relative GPT-4 score and 4\% and 9\% on open and closed set VQA\footnote{Our code, data, and model is publicly accessible at quilt-llava.github.io.}.

\end{abstract}

\begin{figure}[ht!]
\begin{center}
\includegraphics[width=0.9\linewidth]{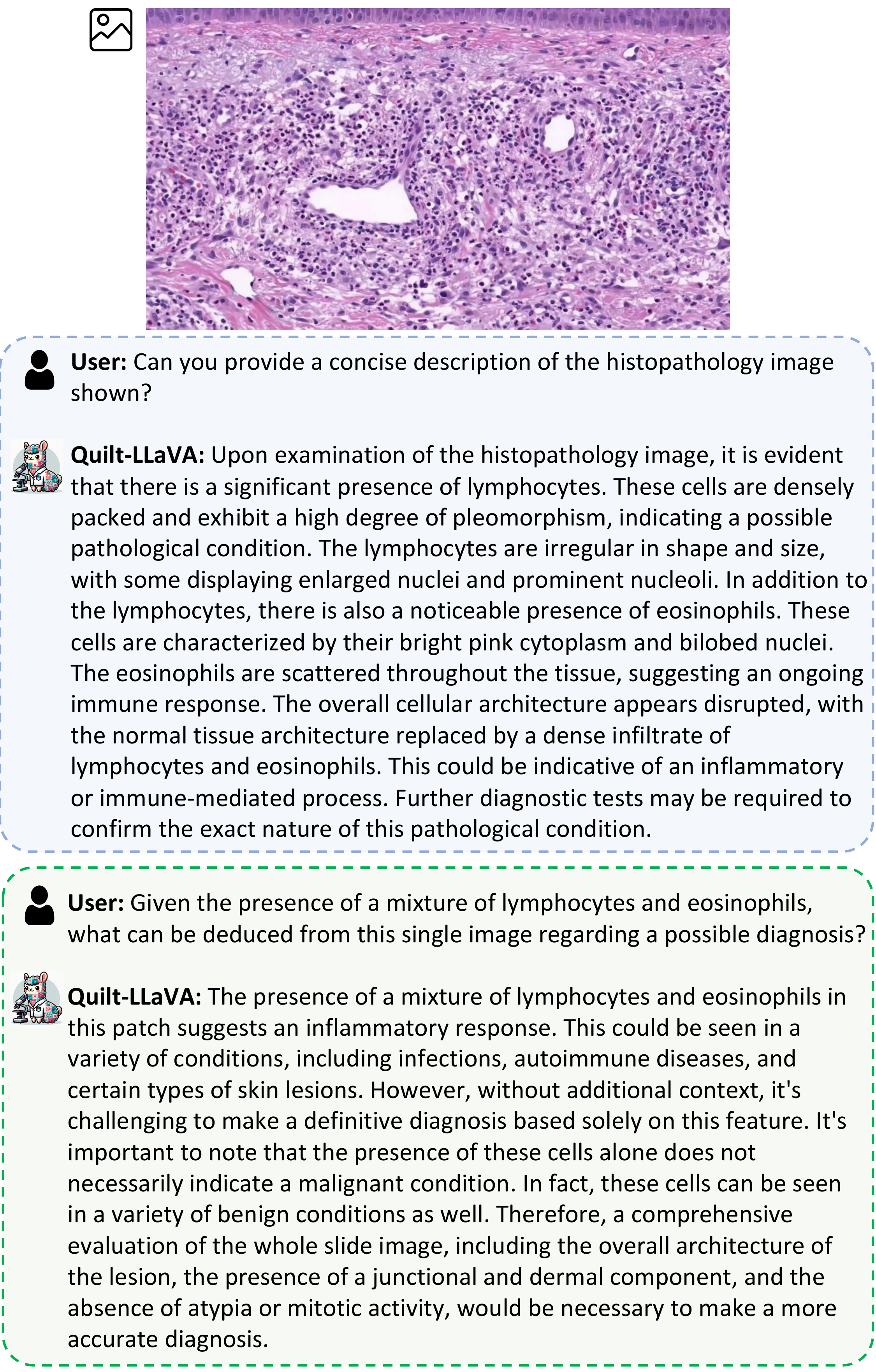}
\end{center}
   \caption{\model is capable of \textcolor{lightblue}{describing} the prominent medical regions within a histopathology patch. Additionally, it can be utilized to \textcolor{lightgreen}{reason} towards a diagnosis based on the current observations. Note: The image includes eosinophils and lymphocytes, and is sampled from a WSI showing rare benign dermatitis.} 
\label{fig:teaser}
\end{figure}

\section{Introduction}
\label{sec:intro}

In histopathology, analyzing giga-pixel-level whole slide images (WSIs) requires a nuanced understanding beyond examining isolated patches taken within the WSI. Pathologists typically analyze these WSIs holistically, compounding evidence from various regions to form a diagnosis. While navigating the WSI regions, pathologists engage in a critical reasoning process over a multi-turn conversation, determining what morphological concepts within the regions confirm or rule out specific diagnoses, making the process dynamic and context-sensitive. For example, in liver tissue WSI, isolated steatosis (fat accumulation) might not be indicative of significant pathology, but if accompanied by ballooning of hepatocytes and lobular inflammation, it strongly suggests non-alcoholic steatohepatitis, a more serious condition that could lead to cirrhosis or liver cancer ~\cite{giashuddin2022histopathological}. This approach involves understanding the diagnostic relevance of histological visual features, spatial relationships, and aggregating insights across the WSI \cite{brunye2023image}. 
Although current histopathology multi-modal models can analyze isolated image patches effectively, they often lack the capability to reason beyond that patch to determine next steps, thereby limiting their utility. 

Meanwhile in natural images, the success of the Large Language and Vision Assistant (LLaVA) \cite{li2023llava} demonstrates the possibility of multi-modal models as chat partners. LLaVA is trained by extracting instruction-tuning data with Large Language Models (LLMs), such as GPT-4~\cite{openai2023gpt}.
Moreover, photos usually include multiple objects which can be referred to using bounding boxes, which can be used to create spatially grounded instruction-tuning data \cite{li2023llava}. Recently, multi-modal models have also begun employing video content, allowing for more complex reasoning beyond single images~\cite{li2023mimic}. 
However, this progress has not translated to histopathology yet, where existing multi-modal models often rely on PubMed articles for image-caption pairs~\cite{li2023llava, moor2023med}. These articles lack two critical elements: \textbf{1)} visually grounded captions necessary for effective visual grounding of histopathology concepts, and \textbf{2)} the broader contextual information from the WSI, essential for reasoning beyond single image patches, limiting their effectiveness in detailed histopathological analysis.

To fill this void, we present \quiltinstruct, an instruction-tuning dataset of $107,131$ histopathology-specific question/answer pairs. Similar to \dataset~\cite{ikezogwo2023quilt}, \quiltinstruct capitalizes on educational histopathology videos from YouTube, featuring pathologists narrating their examination of WSIs. \textbf{1)} To alleviate the limited spatial awareness, we extract narrators' mouse cursors from videos, grounding histopathology concepts within each patch using spatio-temporal clustering. \textbf{2)} To enable reasoning, we propose novel instruction-tuning QA prompting techniques for histopathology: Complex reasoning, and iterative abductive reasoning, which incorporates the global WSI diagnosis and its supporting facts with image captions to ground factual information, preventing hallucinations (see Fig.~\ref{fig:prompts}).

Using \quiltinstruct, we train \model, a multi-modal model for histopathology, with its capabilities illustrated in Fig~\ref{fig:teaser}. 
\model undergoes a two-stage training process (see Fig~\ref{fig:quilt_llava}). First, it is aligned with the histopathology domain using 723K image-text pairs from \dataset \cite{ikezogwo2023quilt}, and then it is further instruction-tuned with \quiltinstruct. \model analyzes given images in detail, spatially localizes medical concepts, and reasons beyond the given image patch by guiding users on what further evidence is needed to validate or rule out certain diagnoses, and can even be used as an educational tool that, instead of directly revealing a diagnosis, subtly hints at it, aiding pathology students in their training (Appendix Section~\ref{sec:training_tool}.)

Finally, educational videos are valuable for their interactivity, as narrators often engage viewers by asking and answering questions throughout their presentations. Leveraging this, we propose \quiltvqa, an organic evaluation dataset extracted from naturally occurring questions and answers from \dataset videos with the help of GPT-4 and manual verification, to evaluate \model's reasoning capabilities. We evaluate \model on \quiltvqa and two public histology VQA test sets on both open and closed questions. 
Using red circle \tikz\draw[red] (0,0) circle (.9ex); \cite{shtedritski2023does} marking the area of interest in the image, we can prompt \model to focus on specific regions of the image patch. We outperform the SOTA by 4\% and 9\% on open and closed set VQA tasks. 
To further evaluate the reasoning capabilities of \model, we utilize GPT-4 to score the model's generated response against two other SOTA multi-modal models: \llava~\cite{liu2023improved} and \llavamed~\cite{li2023llava}. 
\model outperforms \llava and \llavamed by over 16\% and 7\%, respectively, and upon increasing the size of instruction-tuning data and only pre-training for three epochs, we achieve even better results of 10.8\% over \llavamed.

\section{Related work}
\label{sec:related}

We built our work from the expanding body of literature in visual instruction-tuning and video-based dataset generation, with a specific focus on their application in the areas of medical image analysis and histopathology.

\noindent
\textbf{Visual instruction-tuning in natural images.} Thanks to the open-source availability of LLMs ~\cite{touvron2023llama, chiang2023vicuna, jiang2023mistral, taori2023alpaca}, studies in the general vision-language domain have advanced the training of multi-modal models by harnessing implicit visual understanding through generating an instruction-tuning dataset from image captions. Prior work \cite{liu2023visual, zhu2023minigpt, gao2023llama} demonstrated significant capabilities, with LLaVA-1.5 matching GPT-4's performance in certain multi-modal tasks \cite{liu2023improved}.

\noindent
\textbf{Visual instruction-tuning in medical images.} Visual Med-Alpaca \cite{han2023medalpaca} created 54K question-answer pairs for instruction-tuning using GPT-3.5. PMC-VQA \cite{zhang2023pmc} curated a larger multiple choice answer based dataset from general medical domains using PubMed, yet its coverage of histopathology is limited. For instance, LLaVA-Med \cite{li2023llava} employs image captions from PubMed articles and, in cases of brief captions, supplements them with sentences from the article, which may not directly pertain to the referenced figure. Within LLaVA-Med, a subset of 17k images relates to histology, yielding 49K question-answer pairs. Furthermore, these works rely on isolated image-caption pairs for constructing instruction-tuning datasets, which limits GPT-4's capacity to reason beyond its context or, if attempted, raises its likelihood of hallucination.

\noindent
\textbf{Video-based image-text datasets.} Multiple works have attempted to curate datasets from videos in the natural domain ~\cite{zellers2021merlot, zellers2022merlot, voigtlaender2023connecting, li2023mimic, wang2023internvid}. MIMIC-IT \cite{li2023mimic} uses video data to create various prompts, including some reasoning-based ones, to create instruction-tuning datasets. Video Localized Narratives \cite{voigtlaender2023connecting} involve annotators verbally describing an image while concurrently moving their mouse cursor over the regions they refer to, yielding dense visual groundings. Acquiring a dataset in the same way is expensive in histopathology. PathNarratives \cite{zhang2023pathnarratives} employed eight pathologists to manually annotate spatially grounded pathology data; however, it is not available for open access. More recently, \dataset\ \cite{ikezogwo2023quilt} was proposed, which curates large-scale data from educational histopathology content from YouTube to create image-caption pairs, however, it lacks spatially grounded captions. We expand upon \dataset's video content to generate grounded histopathology data. To the best of our knowledge, our work is the first to use videos for creating spatially grounded instruction-tuning datasets in the medical domain.

\section{Curating \quiltinstruct}

To construct a comprehensive histopathology instruction-tuning dataset with visual groundings in WSIs, we harness the rich narrative content of educational YouTube videos.

\noindent
\subsection{Data preparation}

\begin{figure*}[ht!]
\begin{center}
\includegraphics[width=1\linewidth]{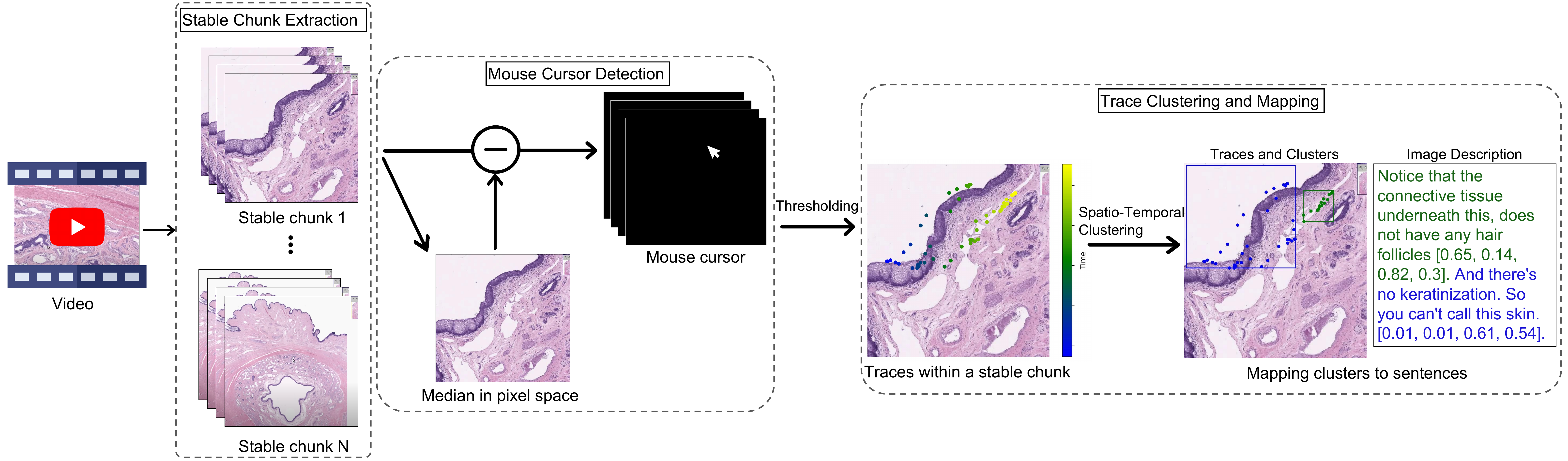}
\end{center}
   \caption{To create \quiltinstruct, we first identify stable chunks within the video. For each chunk, we compute a median frame in the pixel domain and subtract it from every frame within the chunk. We then apply a threshold to reduce noise and take the maximum value to capture the mouse cursor points. These cursor points are then clustered to localize medical content in image captions. Please note that color encodes time in the "Trace Clustering and Mapping" part of the figure.}
\label{fig:pipeline}
\end{figure*}

We make the key observation that, in educational videos, narrators often pause while exploring large-scale WSIs before indicating salient areas with their cursor~\cite{kahneman1973attention, voigtlaender2023connecting, jacko2012human, pont2020connecting}. Our process uses three steps to convert unstructured videos into usable visually-grounded instruction data: First, we \textbf{localize narrators' cursors} in videos. Then, we perform \textbf{spatio-temporal clustering of cursor location} to visually ground histopathology concepts within images. These two steps are illustrated in Fig. \ref{fig:pipeline}. Finally, using the extracted grounded captions, we use an LLM to \textbf{generate our instruction-tuning dataset - \quiltinstruct}. This process involves prompting techniques, from independent prompts generating diverse Q/A pairs for each image patch to reasoning-based prompts combining information across patches in WSIs, creating Q/A pairs that reason towards diagnoses.

\noindent
\textbf{Data source.} We construct \quiltinstruct from 4149 educational YouTube videos totaling over 1,000 hours of content. 
These videos are part of a recent histopathology dataset, \dataset~\cite{ikezogwo2023quilt}.
The videos feature a \textit{narrative style}, where pathologists articulate histopathological concepts while interacting with WSIs. \dataset\ provides rich image-text data but does not offer spatial annotations to connect text descriptions with specific regions in the images—a key aspect often absent in most medical datasets. Fortunately, narrators' mouse cursors highlight morphological features, offering visual grounding to bridge this gap. We next detail the process of extracting these cursors from raw video footage.

\noindent
\textbf{Localizing Cursors}. Extracting the cursor location from histopathology images poses a significant challenge due to the homogeneity in color and texture, alongside the presence of other dynamic elements in videos, such as minor pixel variations and variations in the narrators' cursor movement speed and style. Despite these obstacles, we have made a critical observation: Narrators typically pause their exploration of a WSI before gesturing with their cursor, guiding the audience's attention. Therefore, if the background is stable for several consecutive frames, it is highly likely that the narrator may be using their cursor to mark medically relevant regions. Leveraging this observation, we employ a relatively straightforward method to detect mouse cursors. First, we isolate segments in each video where the background is mostly static, termed stable chunks shown in the leftmost section of Fig. \ref{fig:pipeline}. To detect the stable chunks, we utilize a simple algorithm (Algorithm \ref{algo1} in the Appendix) that detects stable chunks of frames by employing a frame-differencing approach to detect chunks with minimal background movement. Our algorithm computes the absolute difference between consecutive frames and then applies a Gaussian filter for adaptive thresholding to pinpoint frames with minor changes. Given the typically uniform coloration of histopathology backgrounds due to histopathology staining, simple pixel-wise difference yields a high rate of false positives, namely, chunks that had a change occur in them are identified as stable. To alleviate this, we incorporate a perceptual metric, Structural Similarity Index Measure (SSIM). By evaluating SSIM on randomly sampled patches, we further verify whether a change has occurred between two frames. For each stable chunk where the background remains consistent, we calculate a median frame in the pixel domain to represent the background. This median frame is then subtracted from the frames within the stable chunk, and a thresholding is applied to mitigate noise or other artifacts, effectively isolating the cursor as illustrated in the middle section of Fig. \ref{fig:pipeline}. Subsequently, we capture the cursor by identifying the coordinates of the maximum pixel value. However, this approach assumes the absence of other slight movements with respect to pixel changes, which is not always the case, as there may be movements like the narrator's facial expressions within the scene. We address this by applying a face detection model \cite{schroff2015facenet} to mask such distractions, ensuring our analysis concentrates solely on cursor movement. This algorithm offers a surprisingly generalizable way to collect cursor traces from any educational videos with similar {\it narrative styling} at a meager computational cost.

After pinpointing these candidate stable chunks, we retain those with a minimum duration of three seconds. Subsequently, we utilize a histology classifier as described in \cite{ikezogwo2023quilt} to eliminate non-histological segments using each stable chunk's background frame. This methodology generates 132K image-caption pairs, represented as \( \langle I^{v}_{j}, c^{v}_{j} \rangle \), where \( I^{v}_{j} \) is the median frame of the \( j^{th} \) stable chunk within the \( v^{th} \) video, and \( c^{v}_{j} \) is the corresponding caption for \( I^{v}_{j} \), which is obtained by converting the narrator's speech to text similar to the strategy outlined in \cite{ikezogwo2023quilt}. Each pair spans an average of 24 seconds, corresponding to 55 words in the caption. From these 132K image caption pairs, 44,163 exhibits active cursor movement where we preserved cursor position as \( (x^{t}_{j}, y^{t}_{j}) \), where \( x \in [0, W] \) and \( y \in [0, H] \), with \( W \) and \( H \) representing the image width and height, respectively, \( t \) spans from 0 up to the total duration of the \( j^{th} \) stable chunk. Next, we demonstrate our approach for clustering extracted mouse cursors, which serves to visually ground our captions.

\noindent
\textbf{Spatio-temporal clustering of the mouse cursor.} Next we proceed to cluster the tracer points \( (x^{t}_{j}, y^{t}_{j}) \) for each stable chunk to spatially ground narrators' words. For spatio-temporal clustering, we transformed our inputs into \( (x^{t}_{j}, y^{t}_{j}, t, w^{t}_{j}) \), where $t$ is time and $w^{t}_{j}$ denotes the cumulative word count at each cursor position, to encode the pace of speech better, and we have normalized the inputs to ensure each input is scaled within the standardized range of 0 to 1. Given this input format, to achieve a balance between spatial and temporal aspects — specifically, to determine the relative importance of spatial proximity and temporal proximity in cluster formation — we apply an exponential decay to the spatial coordinates, \( e^{-\lambda t} \), where \( \lambda \) is set at 0.05, ensuring that points closer in time are preferentially clustered together. The algorithm dynamically selects the number of clusters based on the number of words in $c^{v}_{j}$. To refine the process of mapping words to clusters, we first determine the temporal center of each cluster. This is achieved by calculating the mean timestamp of the tracer points within a cluster, which gives us a 'temporal midpoint.' Subsequently, we assign words to clusters based on their temporal proximity to these midpoints, ensuring that each word matches the cluster whose average time is closest to the word's occurrence time. Finally, the resulting clusters are represented by bounding boxes, denoted in the format [x1, y1, x2, y2], where each coordinate is a floating-point number normalized between 0 and 1. These coordinates specify the boxes' top-left (x1, y1) and bottom-right (x2, y2) corners. A sample is shown in Fig. \ref{fig:pipeline}. Next, we show how we used our grounded captions to generate our instruction-tuning data \quiltinstruct.

\subsection{Generating \quiltinstruct}

We generate two sets of question-answer types. First, we employ Conversations and Detailed Descriptions prompts, akin to those in \cite{liu2023visual}, that take as input single patch-level grounded-text to generate Q/A pairs constrained by the independent input sample; hence we call these prompts -- Independent prompts. Secondly, we leverage the contextual continuity inherent in our dataset -- that is, for a video reviewing a single WSI, we leverage the sequential unraveling of concepts/clues toward a final diagnosis, by introducing novel Reasoning-based Prompts which receive as input patch-level grounded-text and global WSI-level information. These enable an LLM (GPT-4) to extend its reasoning beyond its immediatecontext while still being anchored by the factual insights derived from the entire video, reducing its hallucinative behavior.

\subsubsection{Independent Prompts}
\label{sec:independent_prompt}

Following the approach of \cite{liu2023visual}, we generate Q/A pairs from each image caption $c^{v}_{j}$, creating an array of questions that aim to explain the context presented in the image. However, these prompts are designed to elicit answers based solely on the information within the caption and do not attempt to extrapolate beyond it.

\noindent
\textbf{Conversation}. The objective of the Conversation prompt is to construct a dialogue between an AI assistant and an individual inquiring about a histopathology image. We generate a diverse set of Q/A pairs based on the image caption $c^{v}_{j}$, which includes bounding boxes of medical concepts. The assistant, equipped with descriptions and bounding box coordinates, responds as if directly observing the image, discussing elements like cell types, activities, and relative positions of medical concepts within the image. The dialogue includes around 3-4 Q/A pairs, maintaining a conversational tone while focusing on the visual content of the images.

\noindent
\textbf{Detailed Descriptions}.
Similar to \cite{liu2023visual}, we instruct GPT-4 to use $c^{v}_{j}$ to generate a detailed description for a given image. Subsequently, questions are randomly selected from a pre-compiled list given in the Appendix Fig.\ref{fig:detailed_description_list}.

\subsubsection{Reasoning-based Prompts}
\label{sec:reasoning_prompt}

Most image-caption pairs used to create instruction-tuning datasets in other works, such as PubMed or COCO \cite{lin2014microsoft}, exhibit contextual isolation, meaning \(<\text{I}_i, \text{c}_i>\) and \(<\text{I}_j, \text{c}_j>\) where \(i \neq j\) do not share context with each other. Consequently, Q/A pairs generated by GPT-4 will be constrained to the context of a single image. Given the gigapixel nature of histopathology images, deriving a comprehensive diagnosis based on a single image patch is often inadequate. The model should ideally reason beyond the given image, guiding the human user on subsequent steps or areas to scrutinize. This may be attempted using GPT-4's inherent medical knowledge. However, coaxing GPT-4 to extrapolate beyond a single caption without an interconnected context could lead to hallucinations, compromising data quality. 

In contrast, our approach capitalizes on video-extracted image-caption pairs derived from a single patient's WSI. To that end, we manually reviewed the entire video content of \dataset\, identifying 2066 videos that feature only a single WSI for our reasoning-based prompts. By utilizing the entire text from these single whole-slide videos, we initially used GPT-4 (see Fig. \ref{fig:diag_prompt} in Appendix) to infer the final diagnosis $d^{v}_j$ and the supporting facts $f^{v}_j$. Utilizing $f^{v}_j$ and $d^{v}_j$ enables us to craft reasoning-based prompts that implicitly direct GPT-4 towards abductive reasoning. This approach fosters extrapolations more anchored in context, reducing hallucination tendency. We developed two prompt types, \textbf{Complex Medical Reasoning} and \textbf{Iterative Abductive Reasoning}, for reasoning-based instruction-tuning dataset generation.

\noindent
\textbf{Complex Medical Reasoning}. Given a caption $c^{v}_{j}$, along with a diagnosis $d^{v}_{j}$ and contributory facts $f^{v}_{j}$, we prompt GPT-4 in a diagnostic reasoning task designed to extrapolate beyond the immediate context of $c^{v}_{j}$. More broadly, we instruct GPT-4 to utilize its inherent medical knowledge to interpret the contents of a single image caption $c^{v}_{j}$, while subconsciously incorporating the diagnosis and supporting facts extracted from the entire video. If the observations from $c^{v}_{j}$ suffice for making a diagnosis based on GPT-4's general medical knowledge, it proceeds to provide a diagnosis along with the facts leading up to it. If not, it extrapolates using $f^{v}_{j}$ to suggest what to look for in the WSI to validate or rule out certain diagnoses. We term this approach "Subconscious Knowledge Injection," as it allows GPT-4 to leverage its medical knowledge while being subtly guided by the provided $d^{v}_{j}$ and $f^{v}_{j}$, effectively constraining its context and ensuring focused reasoning and fewer hallucinations.

\noindent
\textbf{Iterative Abductive Reasoning}: We simulate a conversation between two GPT-4 agents, mimicking a scenario where a professional pathologist uses our model to ask longer medically intricate questions about an image. This contrasts with the shorter questions typically found in other prompts that we use. The first agent, termed Human-GPT, is provided with an image patch caption $c^{v}_{j}$ and is tasked with abductively reasoning about the possible diagnoses and the facts used to arrive at these conclusions. This is presented in the format: \textit{User: [\{Abduction: xxx\}, \{Facts Used: xxx\}]}. The second agent, referred to as the AI Assistant GPT, is privy to the diagnosis $d^{v}_{j}$ and facts $f^{v}_{j}$, simulating someone who has viewed the WSI of this particular patient. The AI Assistant evaluates the accuracy of the abduction derived by Human-GPT and provides comments or hints at potentially overlooked details using its inherent medical knowledge while utilizing $d^{v}_{j}$ and $f^{v}_{j}$. This is communicated in the format: \textit{GPT: [\{Comments: xxx\}, \{Hint: xxx\}]}. If Human-GPT has exhausted all information from $c^{v}_{j}$ and a diagnosis cannot be made, the AI Assistant suggests exploring other patches for additional evidence, such as "Consider looking for evidence of X in other patches to validate your diagnosis.". Furthermore, leveraging Iterative Abductive Reasoning-based instruction-tuning data, \model can serve as an educational tool, aiding human users in brainstorming by providing hints about the next steps, without fully revealing the answer. (See Fig. \ref{fig:educational_reasoning} in Appendix). The conversation between the AI Assistant and Human-GPT proceeds with an upper limit of iterations, randomly selected between two to four interchanges per agent, where after each exchange, the start of a new exchange incorporates the post-exchange history, allowing GPT to retain the memory of past conversations. The dialogue may terminate before reaching this limit if Human-GPT arrives at a conclusive diagnosis, or if the AI Assistant determines that Human-GPT has fully exhausted all relevant information from $c^{v}_{j}$, and then guides it on what to do next using $f^{v}_{j}$.

\noindent
\textbf{Dataset Statistics.} We extracted 162,566 image-caption pairs from \dataset. To refine this data, we filtered out captions with fewer than 20 words and those with more than 150 words. This process resulted in a dataset of 114,343 pairs, with an average caption length of 55 words. From this, we created \quiltinstruct, comprising 107,131 question/answer pairs where, on average, we have questions with 16.5 words and answers with 101 words. For reasoning-based prompts, we manually reviewed 4,149 videos and selected 2,066 that focused on a single WSI from a single patient.

\begin{figure*}[!t]
\begin{center}
\includegraphics[width=1\linewidth]{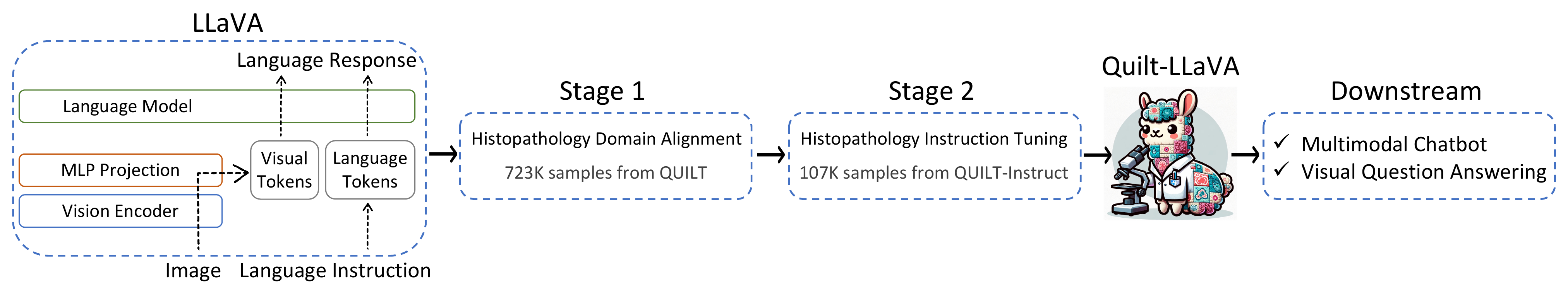}
\end{center}
   \caption{\model was initialized with the general-domain LLaVA and trained for two stages: Histopathology Domain Alignment on \dataset and instruction-tuning on \quiltinstruct. We evaluated \model on visual conversation and question answering tasks.}
\label{fig:quilt_llava}
\end{figure*}

\section{Training \model \& evaluating with \quiltvqa}

In this section, we detail using \quiltinstruct\ to train \model. Next, we curate \quiltvqa independently of \quiltinstruct, to evaluate \model. Finally, we generate an Instruction Following Test Set from \quiltvqa to assess \model's ability to follow instructions.

\subsection{Training \model}

We embrace the \llava autoregressive model architecture for its simple yet efficient design. Additionally, our selection of the \llava architecture aimed to ensure consistency in evaluation against our baselines \llava \cite{liu2023visual} and LLaVA-MED \cite{li2023llava}, both of which utilize a curriculum-learning strategy \cite{li2023llava} on instruction tuning sets. Overall, \llava integrates a vision module, an LLM, and an MLP connector, allowing the LLM to process visual information. Initially, the MLP —serving as a projector— is trained until it converges. During this stage, both LLM and the vision module kept frozen. Subsequently, both the MLP and the LLM are fine-tuned with instruction-following data to align the model with human pathologists. Our overall architecture is shown in Fig.~\ref{fig:quilt_llava}. \llava typically uses a pre-trained CLIP image encoder; for our domain, we use the pre-trained CLIP model trained using public histopathology datasets such as \quiltmodel~\cite{ikezogwo2023quilt} and PLIP ~\cite{huang2023visual}. 
We also run ablations with various image encoders, training strategies, and visual prompts.

\noindent
\textbf{Histopathology domain alignment.} First, we align our vision and language models within the histopathology domain. To that end, we extract 723K image-text pairs from \dataset and convert the captions into the Q/A format. To do that, we randomly select an instruction (Question) and prepend it to the caption (Answer) to create Q/A pairs. The instructions, drawn from a predefined list (See Appendix Fig. \ref{fig:question_list_stage1}), are designed to variably describe the visual information present in the images. At this stage, we freeze vision and language models and only train the MLP layer, which is tasked to project the embeddings from the image encoder to enable the language model predict the images' caption given the question. This stage aligns the histology image embeddings to their corresponding text embeddings.

\noindent
\textbf{Histopathology instruction-tuning.} Finally, we fine-tune our model with \quiltinstruct. At this stage, we keep the visual encoder weights frozen and continue to train the MLP layer and the language module.

\subsection{Evaluation Data Generation: \bf{\quiltvqa}}
\label{sec:eval_data_gen}

In the field of histopathology, researchers rely on evaluation datasets like PathVQA \cite{he2020pathvqa} and PMC-VQA \cite{zhang2023pmc} to assess the performance of their models. However, these datasets exhibit notable shortcomings, including significant repetitiveness due to paraphrasing the same question. Even worse, oftentimes, there are contradictory answers to the same question (See Appendix Section \ref{sec:vqa_issues}). In contrast, educational video content provides a valuable resource: the interactive element introduced by narrators who often pose questions during their presentations and subsequently provide the answers themselves. For instance, a narrator says "\textit{Do you know what kind of organ we’re dealing with?}" and then proceeds to elaborate by "\textit{Yes, this is a colon.}". This Q/A format within the videos offers a rich organic Q/A dataset to be extracted and repurposed for evaluation.

To harness this potential, we map the "?"s in the video's transcript to our stable chunks. If a "?" appears within a 45-second time frame of a stable chunk, we expand the stable chunk's caption to encompass the complete sentence that includes the question mark. This method ensures that the questions are related to the visual content displayed. Following the data pre-processing and the strategic mapping of question marks, we prompt GPT-4 to extract the question and answer pairs directly given in a text. We input GPT-4 the stable chunk's text, as well as any sentences from the text that end in a question mark, indicating a posed question. Fig. \ref{fig:eval_prompt} in the Appendix demonstrates our GPT-4 prompt to generate \quiltvqa. After the initial extraction by GPT-4, we perform a manual verification, ensuring that each question-answer pair in our dataset is not only medically relevant but also correctly corresponds to the content provided in the stable chunk's text. See table \ref{tab:data_stat_vqa_datasets} in the Appendix for statistics of \quiltvqa. 
Additionally, we categorize questions into two groups: Image-dependent (1055 Q/A pairs), referencing the narrator's specific image, and general-knowledge (228 Q/A pairs), rooted in broader medical understanding. This division lets us craft evaluation sets that comprehensively test the model's image analysis and medical knowledge.

\subsection{Evaluation data generation: \textbf{Instruction Following Test Set}}
\label{sec:chat_eval_data_gen}

In addition to \quiltvqa, which is focused on assessing \model's medical knowledge, we also aimed to evaluate the model's capability in following instructions during multi-modal conversations. To that end, we constructed a set of 326 questions, including 256 conversational and 70 detailed description questions, all derived from image-text pairs extracted from unseen videos within \quiltvqa. To generate this evaluation set, we followed the same Conversation and Detailed Description based prompts that we used when generating \quiltinstruct.

\section{Experiments}

This section covers the performance of \model against existing SOTA multi-modal models on histopathology VQA benchmarks using a variety of metrics. First, using GPT-4 we conducted an alignment of output generations with the ground truth answer. Second, we run open and closed set VQA tasks. Finally, we ablate with visual promptings as well as different trained models.

\begin{table*}[ht!]  
\centering  

\begin{subtable}{1.0\textwidth} 
\centering
\tablestyle{3pt}{1.2}
\begin{tabular}{l|cc|cccccccccccccc|c}  
\toprule
&  \multicolumn{2}{c|}{\textbf{Question Types}}  & \multicolumn{14}{c|}{\bf Sub-Domains}  & {\bf Overall}    \\
&   Conv &  Desc & Bone & Breast & Cyto & Derm & Endo &  Gastro & Geni & Gyne & H\&N & Hema & Neuro & Pulm & Renal & Soft \\
(Question Count)  & (256) & (70) & (25) & (23) & (23) & (21) & (23) & (23) & (23) & (22) & (24) & (22) & (24) & (25) & (23) & (25) & (326)  \\ \midrule
 \llava ~\cite{liu2023improved}  & 61.4 & 36.5 & 54.5 & 62.0 &  49.2 & 48.0 & 60.1 & 49.5 &  62.5 & 62.2 & 61.9 & 49.7 & 59.7 & 44.8 & 53.9 & 62.7 & 55.7 \\
LLaVA-MED ~\cite{li2023llava} & 70.1 & 46.9 & 62.1 & 69.3 &  54.1 & 64.0 & 61.0 & 60.7 &  71.2 & 68.1 & 70.3 & 66.9 &  66.0 & 58.9 & 62.7 & 73.4 & 64.8   \\ \hline
\model @  \halfinstruct & 76.3 & 58.7 & 83.4 & 73.3 &  69.2 & 66.7 & 71.7 & 67.2 &  84.5 & 81.1 & 78.4 & 63.2 &  68.9 & 55.2 & 63.5 & 87.7 & 72.3   \\
  \rowcolor{gray!50}
 \model @ \fullinstruct & 78.4 & 66.0  & 82.5 & 84.4 & 75.0 & 79.0 &  76.2 & 72.8 & 75.3 & 82.1 &  79.1 & 69.1 & 68.7 & 58.1 &  67.8 & 89.0 & 75.6 \\ 
\end{tabular}  
\vspace{1mm}
\caption{Performance comparison of multi-modal chat instruction-following abilities, measured by the relative score via language GPT-4 evaluation. Our best model \model with ViT-B-32 Vision Encoder~\cite{ikezogwo2023quilt}, 7B Language Model (trained for Stage1: 3 epochs, Stage2:1 epoch) outperforms the baselines.}
\label{tab:chat_perf}
\end{subtable}  


\begin{subtable}{1.0\textwidth} 
    \centering 
    \tablestyle{3pt}{1.2}
    \begin{tabular}{lcc|rr|c|rr|rr|rr}  
    \multicolumn{3}{c|}{\bf \model Model Variants}    & \multicolumn{2}{c|}{\bf PathVQA}      & \multicolumn{1}{c|}{\bf PMC-VQA-Subset}    & \multicolumn{2}{c|}{\bf \quiltvqa}     &   \multicolumn{2}{c|}{\bf \quiltvqa \tikz\draw[red] (0,0) circle (.9ex);}     & \multicolumn{2}{c}{\bf Average}                  \\
    Instruct &   Stage 1 & Stage 2  & Open    & Closed  & Closed  & Open   & Closed  & Open (w/o \tikz\draw[red, fill=red] (0,0) circle (.5ex); w/)   & Closed (w/o \tikz\draw[red, fill=red] (0,0) circle (.5ex); w/)   & Open   & Closed   \\ 
      \shline
      \multicolumn{10}{l}{\it \quiltmodel ViT-B-32 Vision Encoder~\cite{ikezogwo2023quilt}, 7B Language Model} \\  
    \fullinstruct & 1 & 0  & 14.34 & 53.78 & 27.05 & 47.69 & 56.56 &  49.62 \tikz\draw[red, fill=red] (0,0) circle (.5ex);  54.13  & 55.56 \tikz\draw[red, fill=red] (0,0) circle (.5ex); 58.33 &  41.45 & 50.26 \\

    \fullinstruct & 1& 1  & 14.24  & 58.42 & 19.63 & 59.82 & 64.43 &  58.81 \tikz\draw[red, fill=red] (0,0) circle (.5ex); 61.08   & 68.52 \tikz\draw[red, fill=red] (0,0) circle (.5ex); 70.37  & 48.49   &   56.27  \\
    
    \fullinstruct & 1& 3  & 12.79 & 56.30 & 17.21 & 57.62 & 63.55 & 56.21 \tikz\draw[red, fill=red] (0,0) circle (.5ex); 58.32  & 65.74 \tikz\draw[red, fill=red] (0,0) circle (.5ex); 69.44 &  46.24   &  54.45   \\
    
    \fullinstruct & 3& 1  & \color{red} 15.30 & 54.93 & 16.01 & \color{red} 60.97 & 60.64 & \color{red} 59.24 \tikz\draw[red, fill=red] (0,0) circle (.5ex); \color{red} 64.06  & 56.48 \tikz\draw[red, fill=red] (0,0) circle (.5ex); 59.26  & \color{red} 49.89   & \color{black} 49.46    \\
    
\multicolumn{10}{l}{\it LLAVA ~\cite{liu2023improved} checkpoint, 7B Language Model } \\ 
    \fullinstruct & 1 & 0  & 11.65  & 54.03 & 33.91 & 55.80 & 58.02 &  54.77 \tikz\draw[red, fill=red] (0,0) circle (.5ex); 59.56  &  51.85 \tikz\draw[red, fill=red] (0,0) circle (.5ex); \color{black}  60.19 &  45.45 & 51.60 \\
    \rowcolor{gray!50} \fullinstruct & 1 & 1  & 15.06  & \color{red} 58.68 & 28.56 & 55.39 & \color{red} 68.81 &  54.24 \tikz\draw[red, fill=red] (0,0) circle (.5ex); 59.83  & \color{red} 71.30 \tikz\draw[red, fill=red] (0,0) circle (.5ex); \color{black} \color{red} 75.00  &  46.13   &   \color{red}60.47  \\
    \hline
    \multicolumn{10}{l}{\it Baselines} \\  
    LLaVA-Med ~\cite{li2023llava} 7B & 0  & 0 & 11.97 & 56.15 & 1.34 &  54.81 & 61.22 & 52.58 \tikz\draw[red, fill=red] (0,0) circle (.5ex); 53.97  & 69.44 \tikz\draw[red, fill=red] (0,0) circle (.5ex); 64.81  &   43.33  &   50.59  \\
    
    LLaVA ~\cite{liu2023improved} 7B   & 0 & 0 & 11.65 & 54.02 & \color{red} 33.91 & 55.81 & 57.73 & 54.74 \tikz\draw[red, fill=red] (0,0) circle (.5ex); 59.96 & 51.85 \tikz\draw[red, fill=red] (0,0) circle (.5ex); 60.19  &  45.54  &   51.54  \\
    \hline
\end{tabular}  
\caption{Results with varying training epochs at different stages and models alongside baselines. 107K indicates the size of instruct data used in Stage-2.}  
\label{tab:model_ablation_vqa_result} 
\end{subtable}  
 
\begin{subtable}{1.0\textwidth} 
\centering 
\tablestyle{3pt}{1.2}
\begin{tabular}{cccc|rr|c|rr|rr}  
    \multicolumn{4}{c|}{\bf \model Instruction Data}    & \multicolumn{2}{c|}{\bf PathVQA}      & \multicolumn{1}{c|}{\bf PMC-VQA-Subset}    & \multicolumn{2}{c|}{\bf \quiltvqa}     &   \multicolumn{2}{c}{\bf \quiltvqa \tikz\draw[red] (0,0) circle (.9ex);}                     \\
    Conv &   Detail & Complex & Abductive  & Open & Closed  & Closed  & Open   & Closed  & Open (w/o \tikz\draw[red, fill=red] (0,0) circle (.5ex); w/)   & Closed (w/o \tikz\draw[red, fill=red] (0,0) circle (.5ex); w/)     \\ 
      \shline
    10k & 10k & 0  & 0  & 12.56 & 50.99 & 12.04 & 57.52 &  64.14 & 55.81 \tikz\draw[red, fill=red] (0,0) circle (.5ex); 55.29 & \color{red} 68.52 \tikz\draw[red, fill=red] (0,0) circle (.5ex); \color{black} 59.26    \\
    0 & 0 & 10k  & 10k  & \color{red} 17.53 & 44.50 & \color{red} 30.46 & \color{red} 73.95 &  39.65 & \color{red} 74.37 \tikz\draw[red, fill=red] (0,0) circle (.5ex); 69.64 &  39.81 \tikz\draw[red, fill=red] (0,0) circle (.5ex); 41.67    \\
    5k & 5k & 5k  & 5k  & 14.04 & 57.12 & 22.13 & 61.75 &  \color{red}65.60 & 60.15 \tikz\draw[red, fill=red] (0,0) circle (.5ex); 62.04 &  59.82 \tikz\draw[red, fill=red] (0,0) circle (.5ex); \color{red} 67.59    \\
    10k & 10k & 10k  & 10k  & 14.10 & \color{red}57.33 & 22.17 & 61.42 &  \color{red}65.60 & 59.67 \tikz\draw[red, fill=red] (0,0) circle (.5ex); 58.50 &  66.67 \tikz\draw[red, fill=red] (0,0) circle (.5ex); \color{red} 67.59    \\

\end{tabular}  
\caption{Instruction-tuning Ablation: finetuning \model on the instruction-tuning data subsets with varying number of samples in each Q/A type.}
\label{tab:data_ablation_vqa_result}  
\end{subtable}

\caption{Quantitative results on histopathology VQA datasets. For open-set questions, we report recall for our free-form text generation method in column {\it Open}. For closed-set questions, we report the accuracy in column {\it Closed}. Red indicates the best-performing model.}  
\label{tab:medical_vqa_model_performance} 
\end{table*}

\noindent
\textbf{Oracle (GPT-4) Alignment Evaluation.} To assess the effectiveness of \model in multi-modal conversations, using the Instruction Following test-set (see section \ref{sec:chat_eval_data_gen}), we leverage the GPT-4 (language-only model) to evaluate the {\it helpfulness, relevance, accuracy, and level of details} of the
responses from the two assistants (candidate model and GPT-4), and give an overall score on a scale of 1 to 10 of the answers generated from the two assistants, where a higher score indicates better overall performance, while also providing comprehensive explanation of its evaluation, for us to better understand the models. We then compute the relative score using GPT-4
reference score for normalization \cite{li2023llava, liu2023visual}, as seen in Table \ref{tab:chat_perf} across 14 sub-pathologies and 2 QA types. See Fig. \ref{fig:eval_gpt_score} in the Appendix for the exact prompt used. All \model models outperform baselines, with \model trained on single epoch stage-1 and a balanced subset of 40K instruction-tuning pairs for stage-2 fine-tuning outperforming \llava and \llavamed by over 16\% and 7\% respectively, and upon increasing the size of instruction-tuning data and pre-training for three epochs we achieve even better results of 10.8\% over \llavamed.

\noindent
\textbf{Visual Question Answering.} We evaluated \model on three histopathology VQA datasets, the details of which are provided in Table \ref{tab:data_stat_vqa_datasets} in the Appendix. These datasets feature a mix of open-ended and close-ended Q/A pairs. For closed-set questions, accuracy is used as the metric to determine the proportion of correct answers given by the model. In contrast, for open-set questions, we focus on recall to assess how frequently the model's responses include ground-truth tokens~\cite{li2023llava}. We compare all versions of \model against the medical domain \llavamed, and general domain \llava in Table \ref{tab:model_ablation_vqa_result}. All variants of \model outperform \llava. We see the difference in text encoder initialization from Vicuna \cite{chiang2023vicuna} in the first set of experiments paired with pre-trained open-sourced \quiltmodel models doing better on open-set questions with an average of 4\% over SOTA and the text encoder initialization from \llava doing better on closed-set questions with an average of 9\% over SOTA. Performance of \model on closed-set questions showcases our model's instruction following capacity (in both binary {\it yes/no} and true multi-choice {\it \{A/B/C/D\}} QA unlike \llavamed which underperforms on the latter due to lack of following such multi-choice instructions), whilst outperforming both baselines on PathVQA and \quiltvqa, albeit performance is limited on PMC-VQA-Subset, we believe it stems from the PMC-VQA's issues as discussed in Section \ref{sec:vqa_issues} of the Appendix. \model outperforms both baselines on Open-set evaluation on PathVQA and \quiltvqa, with significant margins. We also run ablations with multiple image encoders and training durations in Stages 1 and 2 summarized in Table \ref{tab:medical_vqa_model_performance_ablation} and qualitative examples in Tables \ref{tab:visual_chat_example1_appendix}-\ref{tab:visual_chat_example4_appendix} in the Appendix.

\noindent
\textbf{VQA with Visual Prompting.} We adopt the visual prompting methodology from \cite{shtedritski2023does} for evaluating our model. This involves utilizing the subset of \quiltvqa with bounding boxes to create ellipses that encapsulate the concepts highlighted by these boxes (See Appendix Fig. \ref{fig:red_circle}). Our \model surpasses the two baselines in both open and closed-set visually prompted VQA, achieving over 4\% in open-set and 10\% in closed-set accuracy. We attribute this enhanced performance to our grounded instruction-tuning dataset, underscoring \model's capability to reason within a region of interest, a critical aspect in histopathology.

\noindent
\textbf{Instruction-tuning Ablation Studies.} To study the impact of \quiltinstruct on performance, we train several models: {\it \textbf{I}} -- Only balanced independent prompts data \ref{sec:independent_prompt} model, {\it \textbf{R}} -- Only balanced reasoning prompts data \ref{sec:reasoning_prompt}, and  {\it \textbf{IR}} -- balanced on all prompt types.
We find that {\it \textbf{I}} model performs better on closed-set questions, whereas {\it \textbf{R}} model outperforms on open-set questions. We argue this is due to the contextual distribution of the questions and answers. {\it \textbf{I}} lends itself well to closed-set questions directly asking option-constrained questions and {\it \textbf{R}} lends itself well to open-set questions because these questions often require reasoning about the diagnosis. {\it \textbf{IR}} improves performance on closed-set questions; however, we see a drop in open-set VQA performance.

\begin{table*}[ht!]
\end{table*}

\section{Conclusion and Limitations}

\noindent
\textbf{Limitations.} Our data, derived from raw video footage, inherently contains noise. This can manifest in many ways: clusters may occasionally be generated incorrectly, or erroneous mouse cursor detection. Additionally, despite our efforts to keep its context limited, GPT-4 still hallucinates, leading to instances where \model also hallucinates. Also, despite being explicitly instructed not to, GPT-4 at times reads from the caption rather than extracting information from the image. We discuss the limitations of \model and \quiltinstruct in Section~\ref{sec:appendix_limitations} of the Appendix.

\noindent
\textbf{Conclusion.} We presented: \textbf{1)} \quiltinstruct, in which we automatically extracted human narrators' mouse movements to spatially ground our concepts and leverage WSI to introduce novel reasoning-based prompts. \textbf{2)} Utilizing \quiltinstruct, we trained \model, a multi-modal model that outperforms existing baselines in both open-ended and close-ended histopathology question answering. \textbf{3)} To further test \model's reasoning capabilities, we curated an evaluation dataset, \quiltvqa, comprised of human-generated question-answer pairs from pedagogical videos. In the future, we plan to collaborate with pathologists to appraise our model and we aim to broaden our focus from histopathology to the wider medical field.

\noindent
\textbf{Acknowledgements.} We thank Microsoft for OpenAI credits, Department of Defense W81XWH-20-1-0798, and National Cancer Institute U01 CA231782, and R01 CA200690.

{
    \small
    \bibliographystyle{ieeenat_fullname}
    \bibliography{main}
}

\clearpage
\setcounter{page}{1}
\maketitlesupplementary

\setcounter{section}{0}
\section{\quiltinstruct and \quiltvqa}

\subsection{Total Cost}
\label{sec:data_cost}
We used GPT-4 to create the \quiltinstruct dataset, spending \$8858 in total where \$6600 is spent on input tokens and \$2204 on output tokens, and \$54 on evaluations. In detail, Complex Reasoning yielded 25,838 Q/A pairs, with an average of 13.16 words per question and 267 words per answer. Detailed Descriptions produce 39,620 samples, each having 11.91 words per question and 130.2 words per answer. For Conversations, we generated 34,242 rows, with an average of 3.51 Q/A pairs per row, totaling 44.74 words for questions and 221 words for answers. Finally, for Iterative Abductive Reasoning, there are 14,642 rows with multiple Q/A pairs, averaging 2.84 Q/A pairs per row, comprising 91.94 words per question and 201 words per answer. Furthermore, we spent \$140 on generating 1283 samples in \quiltvqa with an average of 10.43 words per question and 19.16 words per answer. This results in \$8998 total cost.

\begin{algorithm}[H]
    \caption{Stable Video Chunk Detection Algorithm}
    \label{algo1}
    \scriptsize 
    \begin{algorithmic}[1]
        \Procedure{DetectStaticFrames}{video, starttime, endtime}
        \State video = video[starttime:endtime]
        \State $fixedFrames \gets \emptyset$
        \State $ValidFrames \gets \emptyset$
        \State $prevFrame \gets \text{first frame in } video$
        \For{$frame \in \text{rest of frames in } video$}
        \State $absDiff \gets \text{absolute difference between } frame \text{ and } prevFrame$
        \State $absDiffThresh \gets \text{apply adaptive thresholding to } absDiff$
        \State $meanVal \gets \text{mean value of } absDiffThresh$
        \If{$meanVal < 10$}
        \State $fixedFrames \gets fixedFrames \cup {frame}$
        \Else
        \If{$\text{length of } fixedFrames \geq \text{minimum duration}$}
        \State $subclip \gets \text{extract sub-clip of frames with}$ 
        \State \hspace{\algorithmicindent}$\text{constant background from } fixedFrames$
        \For{$\text{randomly sampled patch} \in \text{each frame of subclip} $}
        \State $SSIMVal \gets \text{calculate SSIM of } patch$
        \If{$SSIMVal > \text{threshold}$}
        \State $ValidFrames \gets ValidFrames \cup {frame}$
        \EndIf
        \EndFor
        \EndIf
        \State $fixedFrames \gets \emptyset$
        \EndIf
        \State $prevFrame \gets frame$
        \EndFor
        \State $staticTimestamps \gets \text{get start \& end times from } ValidFrames$
        \State\textbf{return} $staticTimestamps$
        \EndProcedure
    \end{algorithmic}
\end{algorithm}

\subsection{Prompts}

Figures \ref{fig:prompt_conversational}, \ref{fig:prompt_description}, \ref{fig:prompt_complex}, \ref{fig:prompt_iterative_assistant}, and \ref{fig:prompt_iterative_student} demonstrate the prompts used for generating conversational, detailed description, complex medical reasoning, and iterative abductive reasoning (assistant and student) Q/A pairs in \quiltinstruct, respectively. In addition, Figure \ref{fig:diag_prompt} includes the prompt used to extract the diagnosis and supporting facts from single WSI cases. Figure \ref{fig:eval_prompt} shows the propmt used for generating \quiltvqa, our evaluation set described in \ref{sec:eval_data_gen}. And finally, Figure \ref{fig:eval_gpt_score} shows the prompt used for evaluating the conversational chat abilities of \model using GPT-4.



\section{Use Case and Limitations}
\label{sec:appendix_limitations}

\subsection{Use Case of \model}
\label{sec:training_tool}
Despite our best efforts, \model still produces erroneous information, meaning it cannot be fully trusted as an automated diagnostic tool. Instead, it should serve as a support tool for pathologists or other medical professionals who are familiar with the field, enabling them to correct or revise \quiltmodel's responses. Additionally, since \model is capable of reasoning beyond its immediate context, it could assist medical practitioners in challenging cases where they might be at an impasse. Furthermore, thanks to its Iterative Abductive Reasoning-based Q/A data, \model can function as a brainstorming tool. Rather than providing immediate answers, it can offer hints, which is particularly beneficial for pathology students or junior pathologists looking to learn and develop their skills. An example is given in Fig. \ref{fig:educational_reasoning}.



\subsection{\quiltinstruct and \quiltvqa Limitations}
\noindent
\textbf{Limitations of Hand-Crafted Video Processing Algorithms.} When aligning narrators' speech with images, there can be instances where the narrator discusses something irrelevant or refers to previously observed matters, potentially resulting in unaligned image-text pairs. However, our empirical observations indicate that these occurrences are rare. Also, the images extracted from the video might inherently contain noise, such as artifacts like text, drawings, faces of the narrators', etc., as illustrated in Fig. \ref{fig:image_artifacts}. Additionally, in some cases, the patient's medical history, race, gender, etc., are mentioned in the videos, which may or may not be inferred from the aligned image. Moreover, the clustering algorithm relies on simple rules and thresholds, which may not yield the optimal clustering for some cases. Finally, our dataset currently includes only English, limiting the \model to operating in just a single language.


\vspace{2em}
\noindent
\textbf{Limitations of Generating \quiltinstruct Using LLMs.} As we derive visual instruction tuning data from captions using GPT-4, we inevitably inherit its biases and limitations. Additionally, there are instances where GPT-4 does not fully adhere to commands instructing it to generate Q/A datasets as if it were directly observing the image; instead, it sometimes produces responses that reference the description, generating phrases that state "it is mentioned in the description, etc.". Moreover, despite being provided with spatially grounded captions, GPT-4 may occasionally overlook this information, resulting in data that lacks spatial context. Most critically, GPT-4 is still prone to generating inaccurate information, leading to instances of \model producing incorrect statements or hallucinations (Please see Failure Cases in Section \ref{sec:appendix_evaluation}). Regarding legal matters, our commitment to making our dataset publicly available requires strict adherence to OpenAI's terms. These terms clearly stipulate that the \quiltinstruct is to be utilized exclusively for research purposes.



\noindent
\textbf{Limitations of Generating \quiltvqa Using LLMs.} 
Limitations of \quiltvqa also encompass those mentioned for \quiltinstruct above. In addition, despite padding chunks with question marks to ensure that answers fall within the context, there remain cases where narrators provide answers to these questions far beyond the chunk's timeframe. This can lead to instances where the answers are not adequately captured.



\setcounter{table}{7}
\begin{table}[ht]
    \centering
    \tablestyle{4pt}{1.2}
    \begin{tabular}{p{0.25\columnwidth}|>{\RaggedRight}p{0.7\columnwidth}}
    \textbf{Category} & \textbf{Details and Example} \\ \hline
    Wrong answers & PathVQA includes questions with wrong answers.\\ & E.g., Q: Did early ischemic injury increase eosinophilia? \\ & A: No. (Correct answer is "Yes" according to Figure 1-9 in \cite{change2009cellular}.) \\ \hline
    Inconsistent Q/A pairs & PathVQA has similar questions with inconsistent answers. \\ & E.g., Q: Does early ischemic injury show surface blebs, increase eosinophilia of cytoplasm, and swelling of occasional cells? \\ & A: Yes. (Contradicts with example in "Wrong answers".) \\ \hline
    Ambiguous questions & PathVQA includes ambiguously phrased questions. The question below starts with "What is showing" and the answer provides a diagnosis. The question should have been more clear as "what is the diagnosis of" or "what has the following symptoms?" \\ &E.g., Q: What is showing increased eosinophilia of cytoplasm? \\ &A: early (reversible) ischemic injury. \\ \hline
    Repetitive Q/A pairs & Both PathVQA and PMC-VQA generate multiple Q/A pairs for a single image-text pair, leading to repetition. E.g., two different questions in PathVQA about early ischemic injury. (Refer to examples in "Wrong answers" and "Inconsistent Q/A pairs".)\\ \hline
    General-domain Q/A pairs & Most VQA datasets focus on general-domain biomedical questions. E.g., PMC-VQA is generated from PubMed articles and we extracted a histopathology subset from it for evaluation purposes. In addition, PathVQA includes many non-histopathology images. \\ 
    \end{tabular}
    \caption{Summary and examples of issues in public VQA datasets.}
    \label{tab:vqa_issues}
\end{table}

\subsection{Visual Prompting Details and Limitations}
The ellipses are drawn in red \tikz\draw[red] (0,0) circle (.9ex);, with their size set to 0.05\% of the largest dimension of the image, as recommended by \cite{shtedritski2023does}.

In our visual prompt testing, we used red, which may not stand out against the purple-pink hues common in histopathology images as seen in Figure \ref{fig:red_circle}. For better contrast, future work could explore other colors like green or blue, similar to the inks used by pathologists to mark regions of interest in WSIs.

\setcounter{figure}{16}
\begin{figure}[ht!]
\begin{center}
\includegraphics[width=1\linewidth]{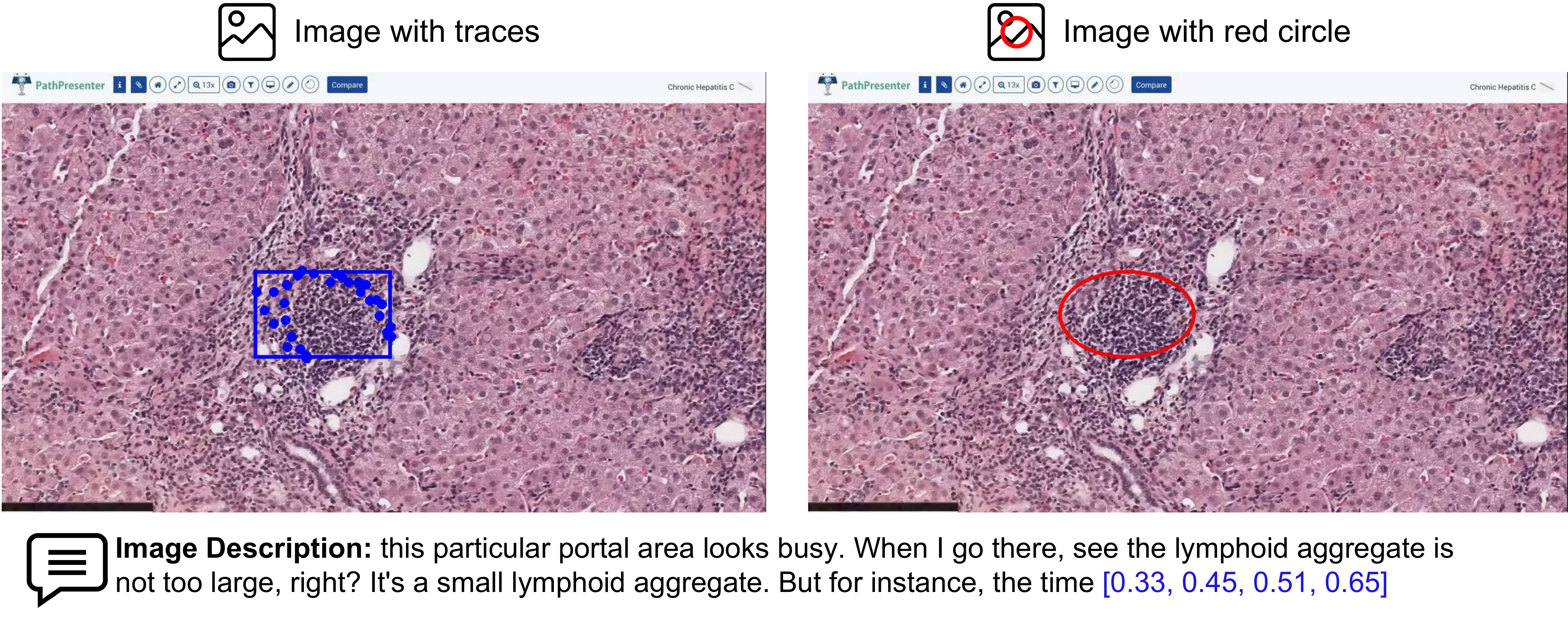}
\end{center}
   \caption{A sample of visual prompting w/ drawing a red circle around the clustered traces.}
\label{fig:red_circle}
\end{figure}

\newpage

\begin{figure*}[ht!]
\begin{center}
\includegraphics[width=1\linewidth]{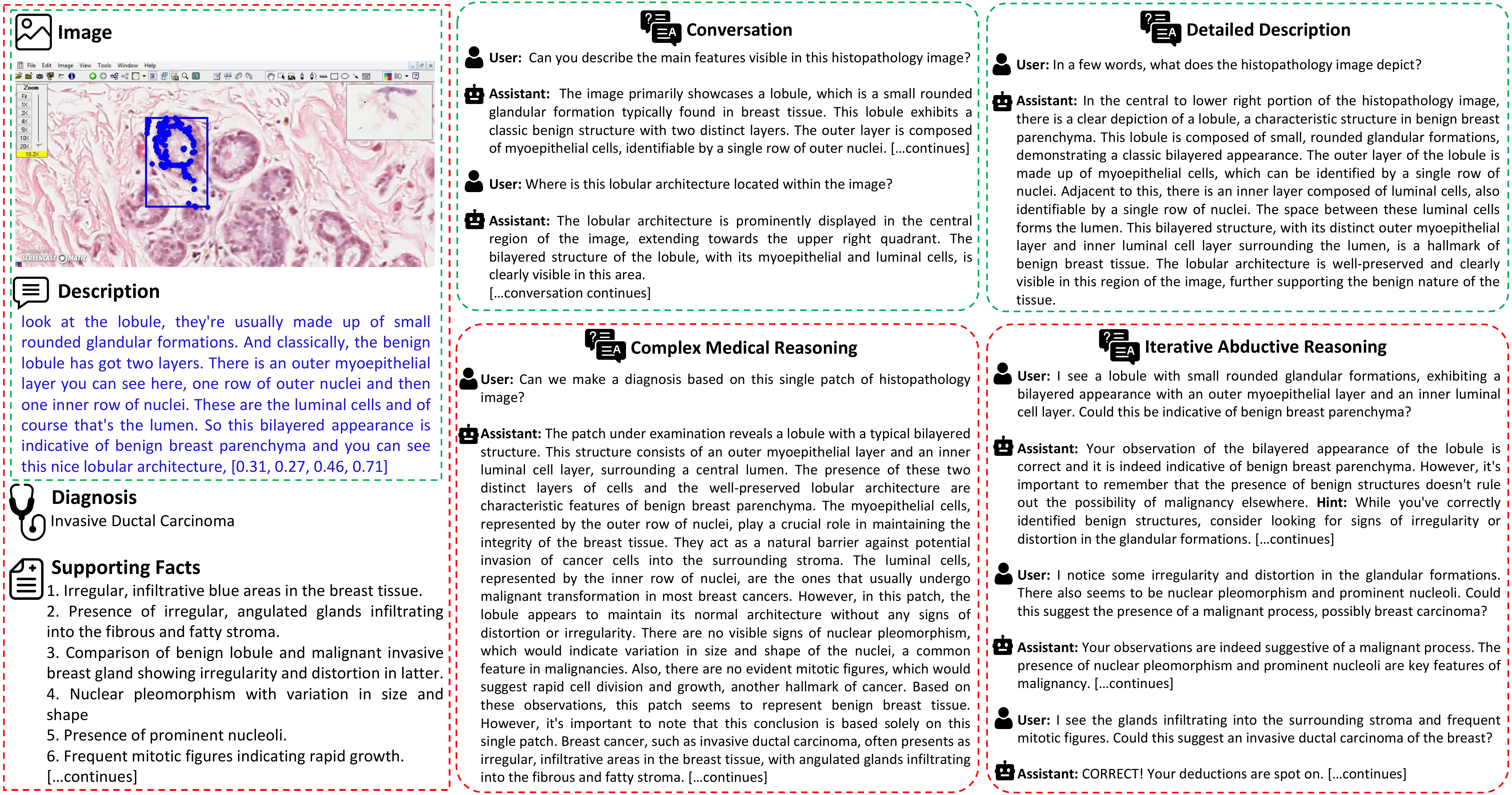}
\end{center}
   \caption{A sample from \quiltinstruct. Only Complex Medical Reasoning and Iterative Abductive Reasoning prompts implicitly utilize Diagnosis and Supporting Facts (which are extracted from the entire video). Note that some of the content is truncated due to space limitations. Please see Figs. \ref{fig:qa_1_2}-\ref{fig:qa_3}-\ref{fig:qa_4} for a more complete overview of \quiltinstruct.}
\label{fig:prompts}
\end{figure*}

\section{Visual Question Answering}

\subsection*{Visual Question Answering Datasets for Histopathology}
\subsection{PathVQA} \cite{he2020pathvqa} is a dataset of 32799 question-answer pairs extracted from 4998 pathology image-caption pairs within textbooks and digital libraries. The questions are categorized as open-ended and closed-ended questions where the former consists of what, where, when, whose, how, how much/how many questions and the latter includes yes/no questions. We use 6761 samples from the evaluation set.

\noindent
\subsection{PMC-VQA} \cite{zhang2023pmc} version-2 includes a test set of 34823 VQA pairs of non-compound images that cover various modalities or diseases. This dataset was curated from image-caption pairs in PMC-OA articles and are in multiple-choice format. We retrieved PMC-VQA-Subset including 2318 histopathology VQA pairs from this dataset.

\noindent
\subsection{Quilt-VQA} consists of 1961 VQA pairs, which we curated from naturally-ocurring Q/A by the narrators of the videos in \dataset. Please see Sec. \ref{sec:eval_data_gen} for details. Figure \ref{fig:quilt_vqa_samples} shows some sample Q/A pairs.

\setcounter{table}{6}
\begin{table}[!h]
\centering 
\tablestyle{4pt}{1.2}
\begin{tabular}{l|c|c|c}  
  & \multicolumn{1}{c|}{\bf PathVQA} & \multicolumn{1}{c|}{\bf PMC-Subset} & \multicolumn{1}{c}{\bf \quiltvqa} \\
\shline
\# Images & 858  &  2209 &  985  \\
 \# QA Pairs & 6761   & 2318   & 1283 \\
 \# Open &   3370  &  -  & 940  \\
 \# Closed &   3391 &  2318 &  343  
\end{tabular}  
\caption{VQA datasets statistics. For PathVQA, we report the evaluation set statistics. PMC-Subset is the histopathology subset retrieved from PMC-VQA.}
\label{tab:data_stat_vqa_datasets}  
\end{table}

\subsection{Visual Question Answering Dataset Issues}
\label{sec:vqa_issues}
Current established VQA datasets have limitations and issues. Table \ref{tab:vqa_issues} summarizes a few of these issues and provides examples from PathVQA \cite{he2020pathvqa} and PMC-VQA \cite{zhang2023pmc} datasets.

\section{Evaluation}

\setcounter{table}{1}
\begin{table*}[h]  
\centering  

\begin{subtable}{1.0\textwidth} 
    \centering 
    \tablestyle{3pt}{1.2}
    \begin{tabular}{lcc|rr|c|rr|rr}  
    \multicolumn{3}{c|}{\bf \model Model Variants}    & \multicolumn{2}{c|}{\bf PathVQA}      & \multicolumn{1}{c|}{\bf PMC-Subset}    & \multicolumn{2}{c|}{\bf \quiltvqa}     &   \multicolumn{2}{c}{\bf \quiltvqa \tikz\draw[red] (0,0) circle (.9ex);}       \\
    Instruct &   Stage 1 & Stage 2  & Open    & Closed  & Closed  & Open   & Closed  & Open (w/o \tikz\draw[red, fill=red] (0,0) circle (.5ex); w/)   & Closed (w/o \tikz\draw[red, fill=red] (0,0) circle (.5ex); w/)     \\ 
      \shline
      \multicolumn{10}{l}{\it \quiltmodel ViT-B-32 Vision Encoder~\cite{ikezogwo2023quilt}, 7B Language Model} \\  
      \fullinstruct & 1 & 0  & 14.34 & 53.78 & 27.05 & 47.69 & 56.56 &  49.62 \tikz\draw[red, fill=red] (0,0) circle (.5ex);  54.13  & 55.56 \tikz\draw[red, fill=red] (0,0) circle (.5ex); 58.33  \\
    \fullinstruct & 1& 1  & 14.24  & 58.42 & 19.63 & 59.82 & 64.43 &  58.81 \tikz\draw[red, fill=red] (0,0) circle (.5ex); 61.08   & 68.52 \tikz\draw[red, fill=red] (0,0) circle (.5ex); 70.37   \\
    \fullinstruct & 1& 3  & 12.79 & 56.30 & 17.21 & 57.62 & 63.55 & 56.21 \tikz\draw[red, fill=red] (0,0) circle (.5ex); 58.32  & 65.74 \tikz\draw[red, fill=red] (0,0) circle (.5ex); 69.44 \\
    \fullinstruct & 3& 1  & 15.30 & 54.93 & 16.01 & \color{red} 60.97 & 60.64 & \color{red} 59.24 \tikz\draw[red, fill=red] (0,0) circle (.5ex); \color{red} 64.06  & 56.48 \tikz\draw[red, fill=red] (0,0) circle (.5ex); 59.26 \\
    \multicolumn{10}{l}{\it \quiltmodel ViT-B-16 Vision Encoder~\cite{ikezogwo2023quilt}, 7B Language Model} \\  
    \fullinstruct & 1 & 1  & 15.14  & 58.24 & 15.83 & 59.61 & 61.52 &  57.46 \tikz\draw[red, fill=red] (0,0) circle (.5ex); 61.04 & 63.89 \tikz\draw[red, fill=red] (0,0) circle (.5ex); 63.89  \\
        \multicolumn{10}{l}{\it CLIP Vision Encoder (ViT-L-14 @336px)~\cite{zhang2023large}, 7B Language Model } \\  
    \fullinstruct & 0 & 1  & \color{red} 15.95  & 59.89 & 15.84 & 58.43 & 69.10 &  56.19 \tikz\draw[red, fill=red] (0,0) circle (.5ex);  59.44  & 70.37 \tikz\draw[red, fill=red] (0,0) circle (.5ex);  62.04 \\
    \fullinstruct & 0 & 3   & 15.29  & \color{red} 61.81 & 17.29 & 55.61 & 68.81 &  54.28 \tikz\draw[red, fill=red] (0,0) circle (.5ex); 57.44   & \color{red}{71.30} \tikz\draw[red, fill=red] (0,0) circle (.5ex); \color{black} 65.74 \\
    \fullinstruct & 1& 1  & 15.19  & 58.54 & 16.18 & 59.01 & 62.10 &  57.01 \tikz\draw[red, fill=red] (0,0) circle (.5ex);  57.98  & 61.11 \tikz\draw[red, fill=red] (0,0) circle (.5ex); 57.41 \\
        \multicolumn{10}{l}{\it LLAVA ~\cite{liu2023improved} checkpoint, 7B Language Model } \\  
    \fullinstruct & 0 & 1  & 14.93 & 56.97 & 27.83 & 55.36 & \color{red} 69.97 & 54.64 \tikz\draw[red, fill=red] (0,0) circle (.5ex); 59.88   &  70.37 \tikz\draw[red, fill=red] (0,0) circle (.5ex); 73.15\\
        \fullinstruct & 0 & 3   & 13.39  & 53.20 & 32.56 & 53.94 & 62.97 & 52.44 \tikz\draw[red, fill=red] (0,0) circle (.5ex);  55.19  &  57.41 \tikz\draw[red, fill=red] (0,0) circle (.5ex); \color{red} 77.78  \\
    \fullinstruct & 1 & 0  & 11.65  & 54.03 & \color{red} 33.91 & 55.80 & 58.02 &  54.77 \tikz\draw[red, fill=red] (0,0) circle (.5ex); 59.56  &  51.85 \tikz\draw[red, fill=red] (0,0) circle (.5ex); \color{black}  60.19 \\
    \fullinstruct & 1 & 1  & 15.06  & 58.68 & 28.56 & 55.39 & 68.81 &  54.24 \tikz\draw[red, fill=red] (0,0) circle (.5ex); 59.83  & \color{red} 71.30 \tikz\draw[red, fill=red] (0,0) circle (.5ex); \color{black}  75.00 \\
    \hline
    \multicolumn{10}{l}{\it Baselines} \\  
    LLaVA-Med ~\cite{li2023llava} 7B & 0  & 0 & 11.97 & 56.15 & 1.34 &  54.81 & 61.22 & 52.58 \tikz\draw[red, fill=red] (0,0) circle (.5ex); 53.97  & 69.44 \tikz\draw[red, fill=red] (0,0) circle (.5ex); 64.81 \\
    LLaVA ~\cite{liu2023improved} 7B   & 0 & 0 & 11.65 & 54.02 & \color{red} 33.91 & 55.81 & 57.73 & 54.74 \tikz\draw[red, fill=red] (0,0) circle (.5ex); 59.96 & 51.85 \tikz\draw[red, fill=red] (0,0) circle (.5ex); 60.19 \\
    \hline
\end{tabular}  
\caption{}  
\label{tab:medical_vqa_model_performance_ablation} 
\end{subtable}
 
\caption{Additional quantitative results on three established histopathology VQA datasets. For open-set questions, we report the recall for our free-form text generation method in column {\it Open}. For closed-set questions, we report the accuracy in column {\it Closed}. Red indicates the best-performing model.}  
\label{tab:medical_vqa_model_performance} 
\end{table*}

\subsection{Qualitative Evaluation}
\label{sec:appendix_evaluation}

\subsection*{Success Cases}

Tables \ref{tab:visual_chat_example1_appendix}, \ref{tab:visual_chat_example2_appendix}, \ref{tab:visual_chat_example3_appendix}, and \ref{tab:visual_chat_example4_appendix} demonstrate successful examples of \model chat abilities compared to \llava and \llavamed.

\setcounter{table}{2}
\begin{table*}[h!]
  \begin{minipage}{0.99\textwidth}
\centering  
\vspace{-4mm}
\scalebox{0.88}{
\begin{tabular}{l p{12.5cm} }
\toprule
 \multicolumn{2}{l}{\bf Visual input example, Histopathology image:}  \\
\midrule
&  \includegraphics[height=3.5cm]{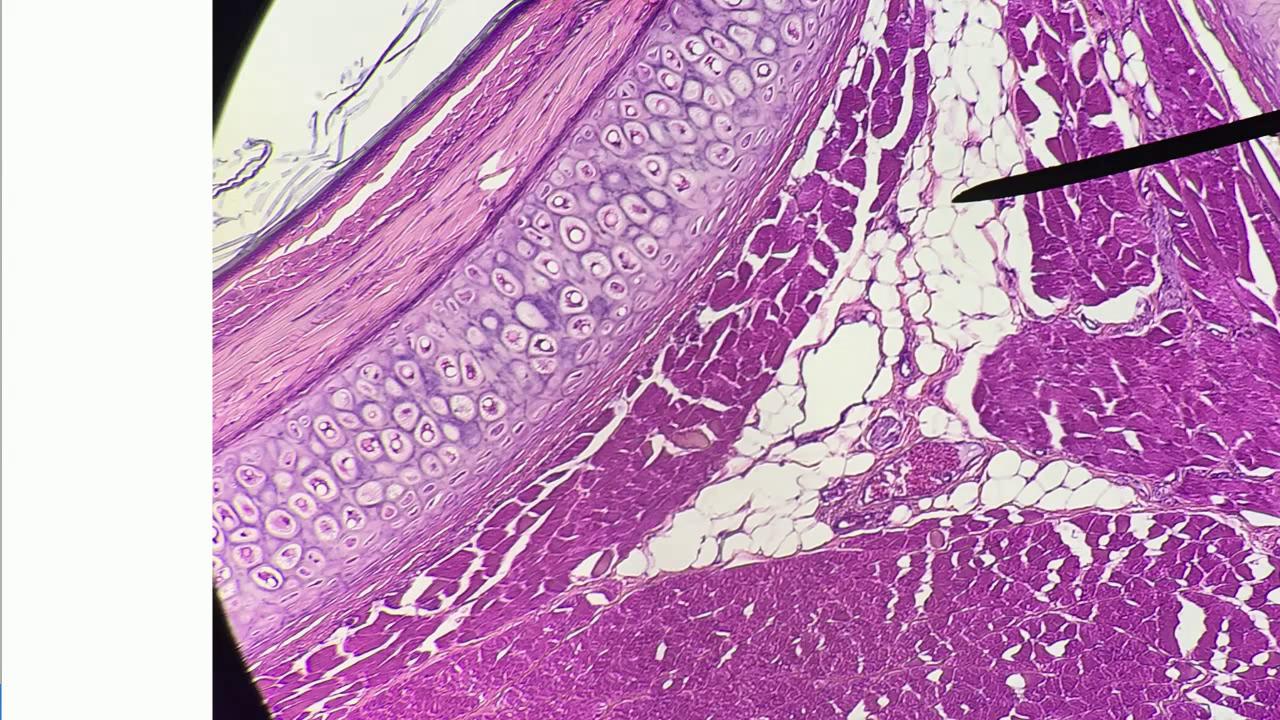} \\
User & What is the connective tissue visible in the image?\\
 \midrule
\model & The connective tissue in the image is identified as the perichondrium. It's a layer of dense irregular connective tissue that surrounds the cartilage. The perichondrium is visible in the upper left quadrant of the image.
\\
\midrule
LLaVA-Med & The connective tissue visible in the image is lamellar keratinized epithelium. This type of tissue is composed of multiple layers of keratinized cells, which are cells that produce keratin, a tough, fibrous protein that provides protection and structural support.
\\
\midrule
LLaVA &  The connective tissue visible in the image is a piece of human muscle tissue.
 \\
 \midrule
Answer &  The connective tissue visible in the image is cartilage.
 \\
\bottomrule
\end{tabular}
}
\vspace{1mm}
\captionof{table}{\color{green} Success case \color{black}: Example comparison of histopathology visual chat and reasoning capabilities. Note: \model correctly understands the histology features and the localization of said features.
}
\vspace{-5mm}
\label{tab:visual_chat_example1_appendix}  
  \end{minipage}
\end{table*}

\setcounter{table}{3}
\begin{table*}
  \begin{minipage}{0.99\textwidth}
\centering  
\vspace{-4mm}
\scalebox{0.88}{
\begin{tabular}{l p{12.5cm} }
\toprule
 \multicolumn{2}{l}{\bf Visual input example, Histopathology image:}  \\
\midrule
&  \includegraphics[height=3.5cm]{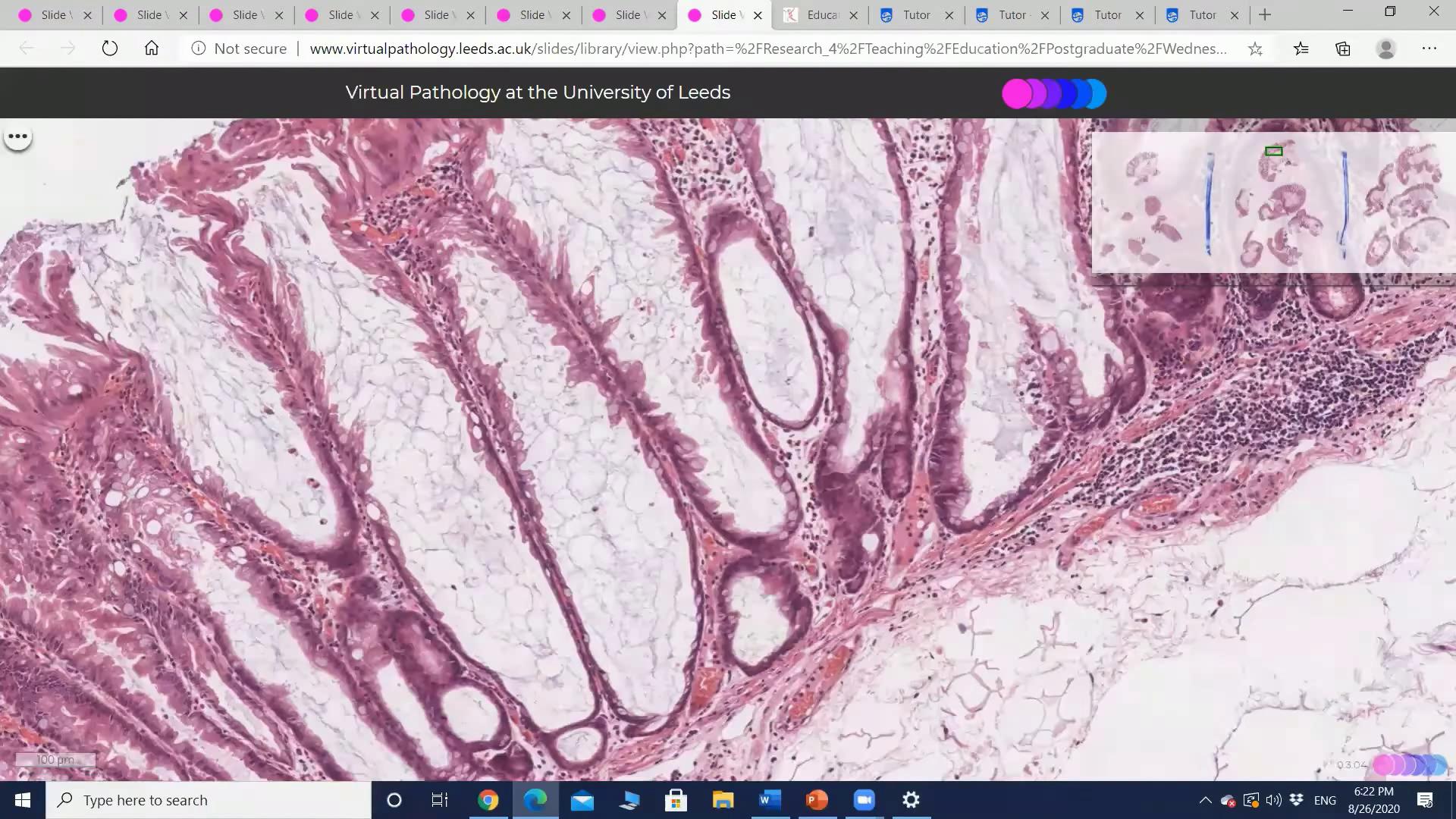} \\
User & What is visible in the middle of the image?\\
 \midrule
\model & In the middle of the image, there is a significant presence of mucin. Mucin is a protein that forms mucus, and its presence in large amounts can be indicative of certain conditions.
\\
\midrule
LLaVA-Med & In the middle of the image, you can see a histopathological section of the human prostate. Histopathology refers to the microscopic examination of tissue to study the manifestations of disease.
\\
\midrule
LLaVA &  In the middle of the image, there is a close-up view of a human organ, specifically the stomach.
 \\
 \midrule
Answer &  Mucus is visible in the middle of the image.
 \\
\bottomrule
\end{tabular}
}
\vspace{1mm}
\captionof{table}{\color{green} Success case \color{black}: Example comparison of histopathology visual chat and reasoning capabilities. Note: \model can correctly answer spatial questions.}
\label{tab:visual_chat_example2_appendix}  
  \end{minipage}
\end{table*}

\setcounter{table}{4}
\begin{table*}
  \begin{minipage}{0.99\textwidth}
\centering  
\scalebox{0.88}{
\begin{tabular}{l p{12.5cm} }
\toprule
 \multicolumn{2}{l}{\bf Visual input example, Histopathology image:}  \\
\midrule
&  \includegraphics[height=3.5cm]{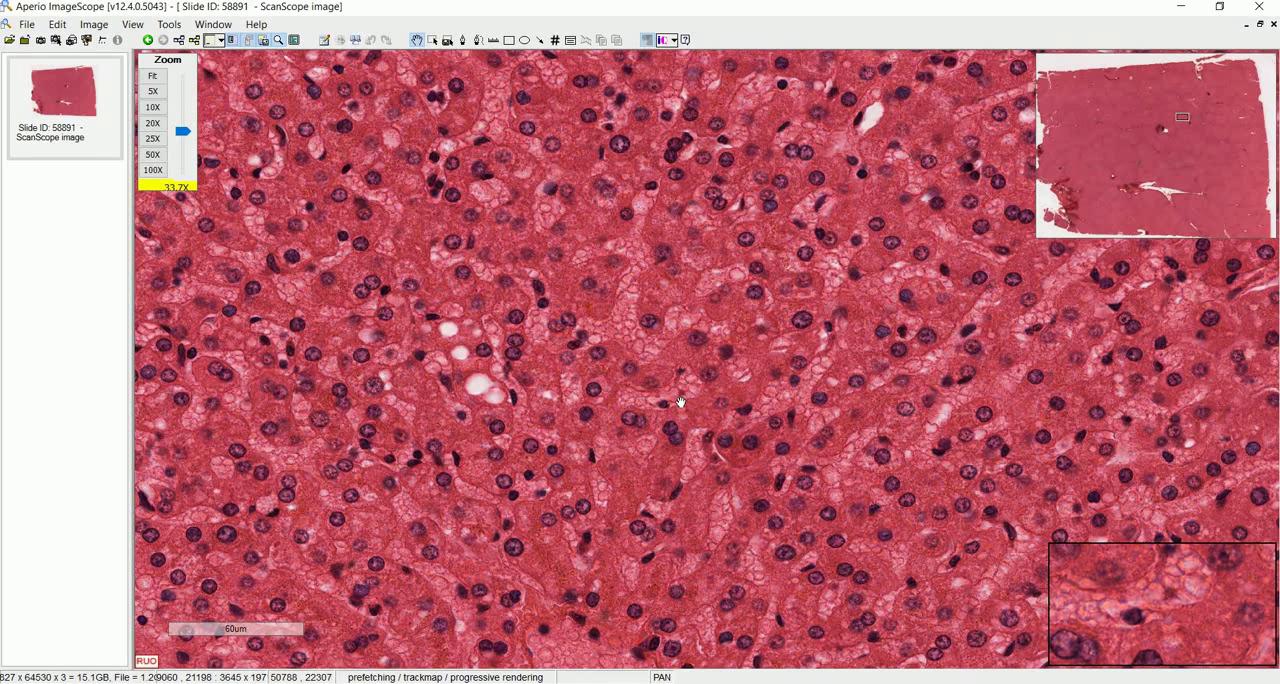} \\
User & What type of cells are the majority in the liver as seen in the image? \\
 \midrule
\model & The image shows that the majority of cells in the liver are hepatocytes, which are the primary cell type in the liver. These cells are located in the central to the right middle region of the image.
\\
\midrule
LLaVA-Med & The majority of the cells in the liver as seen in the image are foamy macrophages.
\\
\midrule
LLaVA &  The majority of cells in the liver as seen in the image are red blood cells.
 \\
 \midrule
Answer &  The majority of the cells in the liver, as seen in the image, are hepatocytes.
 \\
\bottomrule
\end{tabular}
}
\vspace{1mm}
\captionof{table}{\color{green} Success case \color{black}: Example comparison of histopathology visual chat and reasoning capabilities. Note: \model correctly interprets the histologic features in the image and their localization of features.}
\vspace{-5mm}
\label{tab:visual_chat_example3_appendix}  
  \end{minipage}
\end{table*}

\setcounter{table}{5}
\begin{table*}
  \begin{minipage}{0.99\textwidth}
\centering  
\vspace{-4mm}
\scalebox{0.88}{
\begin{tabular}{l p{12.5cm} }
\toprule
 \multicolumn{2}{l}{\bf Visual input example, Histopathology image:}  \\
\midrule
&  \includegraphics[height=3.5cm]{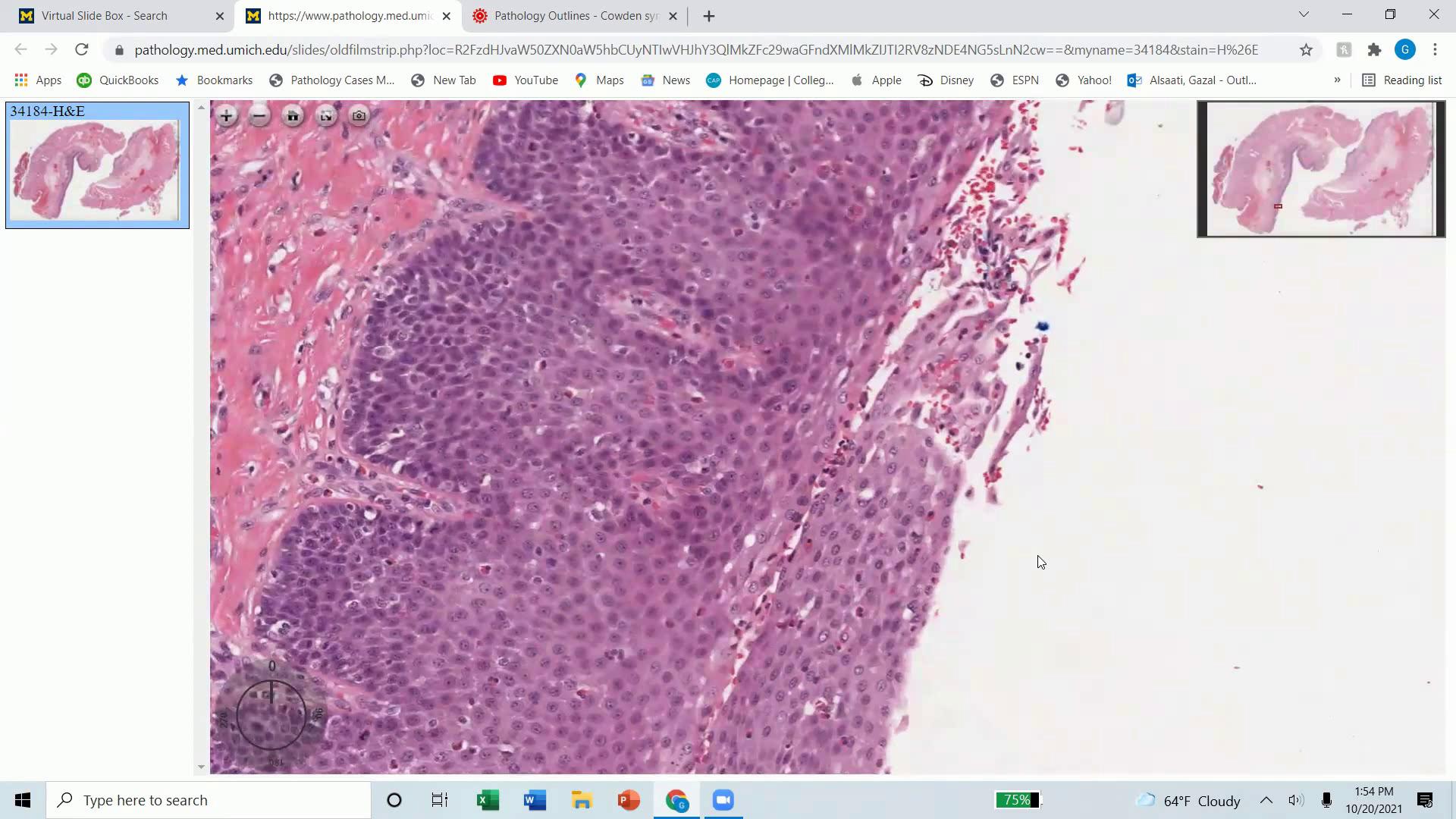} \\
User & What are the three criteria visible in the image that suggest eosinophilic esophagitis?\\
 \midrule
\model & The image shows three key features that are indicative of eosinophilic esophagitis. Firstly, there is a noticeable increase in the number of eosinophils, which are a type of white blood cell. Secondly, there is a significant amount of eosinophilic infiltration, which refers to the accumulation of these cells in the tissue. Lastly, there is evidence of basal cell hyperplasia, which is an increase in the number of cells in the basal layer of the tissue. These three features together strongly suggest a diagnosis of eosinophilic esophagitis.
\\
\midrule
LLaVA-Med & The three criteria visible in the image that suggest eosinophilic esophagitis are: 
1. Increased intraepithelial eosinophils (H\&E, x40)
2. Eosinophilic infiltrates in the lamina propria (H\&E, x40)
3. Presence of eosinophils in the muscularis mucosa (H\&E, x40)

These criteria are based on the histological features of the tissue sample, which can help in the diagnosis of eosinophilic esophagitis.
\\
\midrule
LLaVA &  In the image, there are three criteria visible that suggest eosinophilic esophagitis (EoE). First, the esophageal tissue is inflamed, as indicated by the red and pink coloration. Second, there are numerous eosinophils present in the tissue, which are white blood cells that are typically elevated in EoE cases. Finally, the tissue appears to be thickened, which is another characteristic of EoE. These three criteria together suggest that the image depicts a case of eosinophilic esophagitis.

 \\
 \midrule
Answer &  The three criteria visible in the image that suggest eosinophilic esophagitis are the presence of eosinophils, basal cell hyperplasia, and spongiosis.
 \\
\bottomrule
\end{tabular}
}
\vspace{1mm}
\captionof{table}{\color{green} Success case \color{black}: Example comparison of histopathology visual chat and reasoning capabilities. Note: \model correctly extracts the morphology features to answer the question. }
\vspace{-5mm}
\label{tab:visual_chat_example4_appendix}  
  \end{minipage}
\end{table*}

\subsection*{Failure Cases}
Tables \ref{tab:fails_visual_chat_example1_appendix}, \ref{tab:fails_visual_chat_example2_appendix}, \ref{tab:fails_visual_chat_example3_appendix}, and \ref{tab:fails_visual_chat_example4_appendix} demonstrate successful examples of \model chat abilities compared to \llava and \llavamed across a few Q/A types (e.g observations, staining, localization etc.).

\setcounter{table}{8}
\begin{table*}[h!]
  \begin{minipage}{0.99\textwidth}
\centering  
\vspace{-4mm}
\scalebox{0.88}{
\begin{tabular}{l p{12.5cm} }
\toprule
 \multicolumn{2}{l}{\bf Visual input example, Histopathology image:}  \\
\midrule
&  \includegraphics[height=3.5cm]{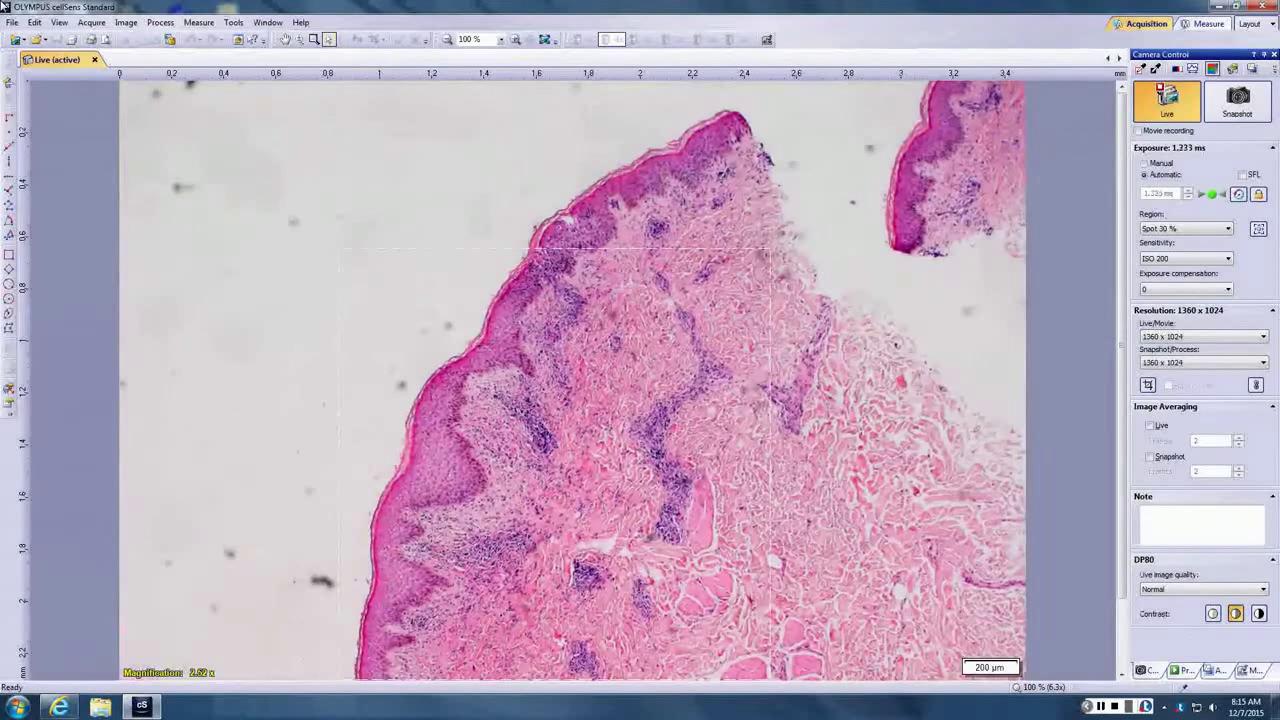} \\
User & What can be observed in the epidermis in the image?\\
 \midrule
\model & The epidermis in the image appears to be quite normal. It's located towards the upper left quadrant of the image, extending towards the center. The epidermis is the outermost layer of the skin and is responsible for providing a barrier against environmental damage.\\
\midrule
LLaVA-Med & In the epidermis of the control group, you can observe the presence of a stratum corneum, which is the outermost layer of the skin. Additionally, there is a keratin layer, which is a protein that provides structural support and protection to the skin.
\\
\midrule
LLaVA &  In the image, the epidermis is observed to be affected by a disease, possibly cancer, as it appears to be infected or affected by a growth.
 \\
 \midrule
Answer &  Spongiosis and erythrocytes can be observed in the epidermis in the image.
 \\
\bottomrule
\end{tabular}
}
\vspace{1mm}
\captionof{table}{\color{red} Failure case \color{black}: Example comparison of histopathology visual chat and reasoning capabilities. }
\vspace{-5mm}
\label{tab:fails_visual_chat_example1_appendix}  
  \end{minipage}
\end{table*}

\setcounter{table}{9}
\begin{table*}[h!]
  \begin{minipage}{0.99\textwidth}
\centering  
\vspace{-4mm}
\scalebox{0.88}{
\begin{tabular}{l p{12.5cm} }
\toprule
 \multicolumn{2}{l}{\bf Visual input example, Histopathology image:}  \\
\midrule
&  \includegraphics[height=3.5cm]{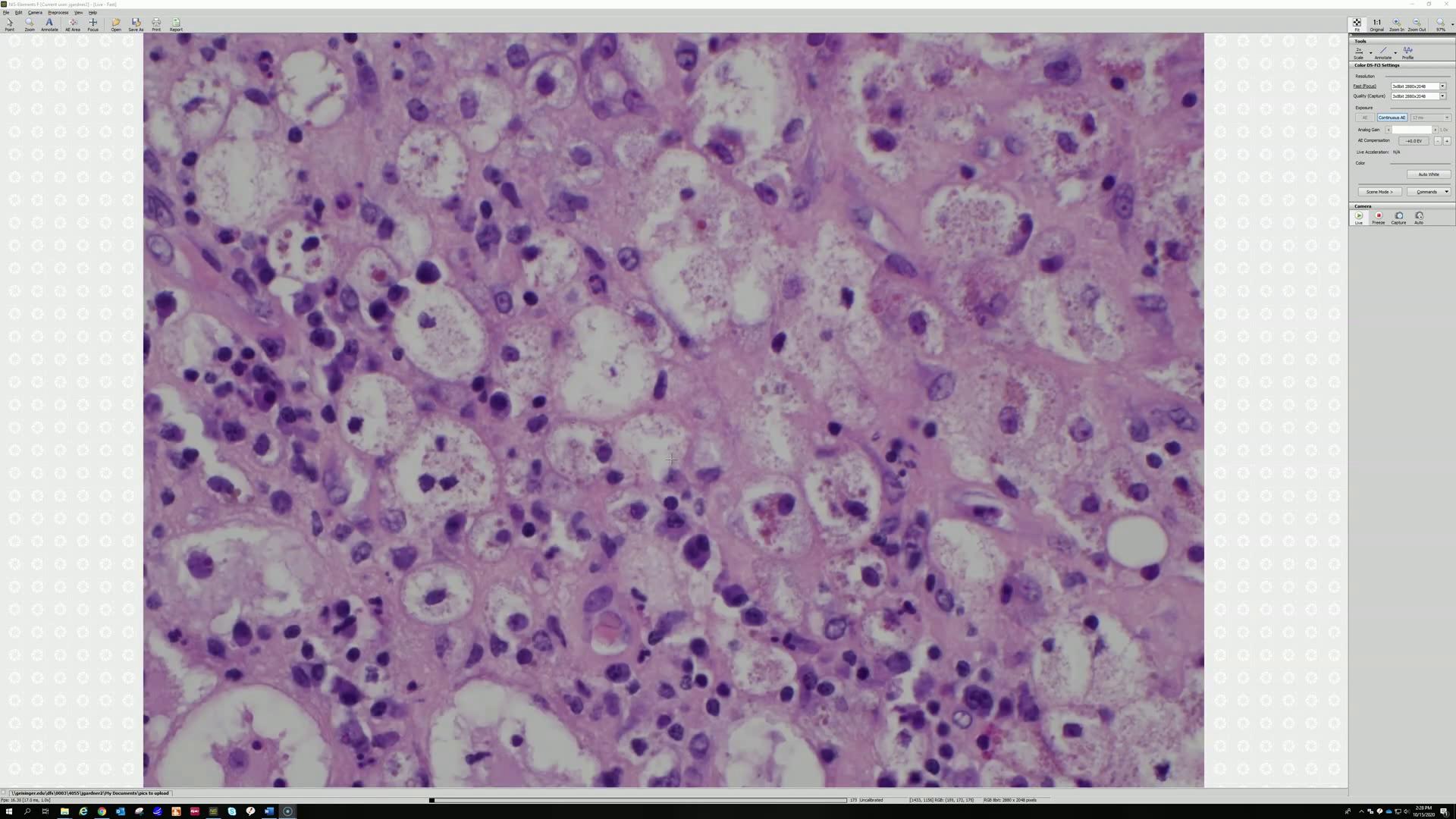} \\
User & What stain can be used to highlight the histiocytes in the image?\\
 \midrule
\model & The histiocytes in this image can be highlighted using a Ziehl-Neelsen stain. This stain is particularly useful for identifying acid-fast bacilli, which are often associated with certain types of infections.\\
\midrule
LLaVA-Med & The image uses hematoxylin and eosin (H\&E) staining, which is a common staining technique used to highlight histiocytes. Histiocytes are a type of immune cell that plays a role in the body's defense against pathogens and foreign substances.
\\
\midrule
LLaVA &  In the image, the histiocytes can be highlighted using a stain called hematoxylin. This stain is commonly used in histology to stain various cell types, including hematopoietic cells, and it helps to differentiate them from other cell types in the tissue. The presence of hematoxylin in the image indicates that the histiocytes are being stained, making them more visible and easier to identify.
 \\
 \midrule
Answer &  The S100 stain can be used to highlight the histiocytes in the image.
 \\
\bottomrule
\end{tabular}
}
\vspace{1mm}
\captionof{table}{\color{red} Failure case \color{black}: Example comparison of histopathology visual chat and reasoning capabilities. }
\vspace{-5mm}
\label{tab:fails_visual_chat_example2_appendix}  
  \end{minipage}
\end{table*}

\setcounter{table}{10}
\begin{table*}[h!]
  \begin{minipage}{0.99\textwidth}
\centering  
\vspace{-4mm}
\scalebox{0.88}{
\begin{tabular}{l p{12.5cm} }
\toprule
 \multicolumn{2}{l}{\bf Visual input example, Histopathology image:}  \\
\midrule
&  \includegraphics[height=3.5cm]{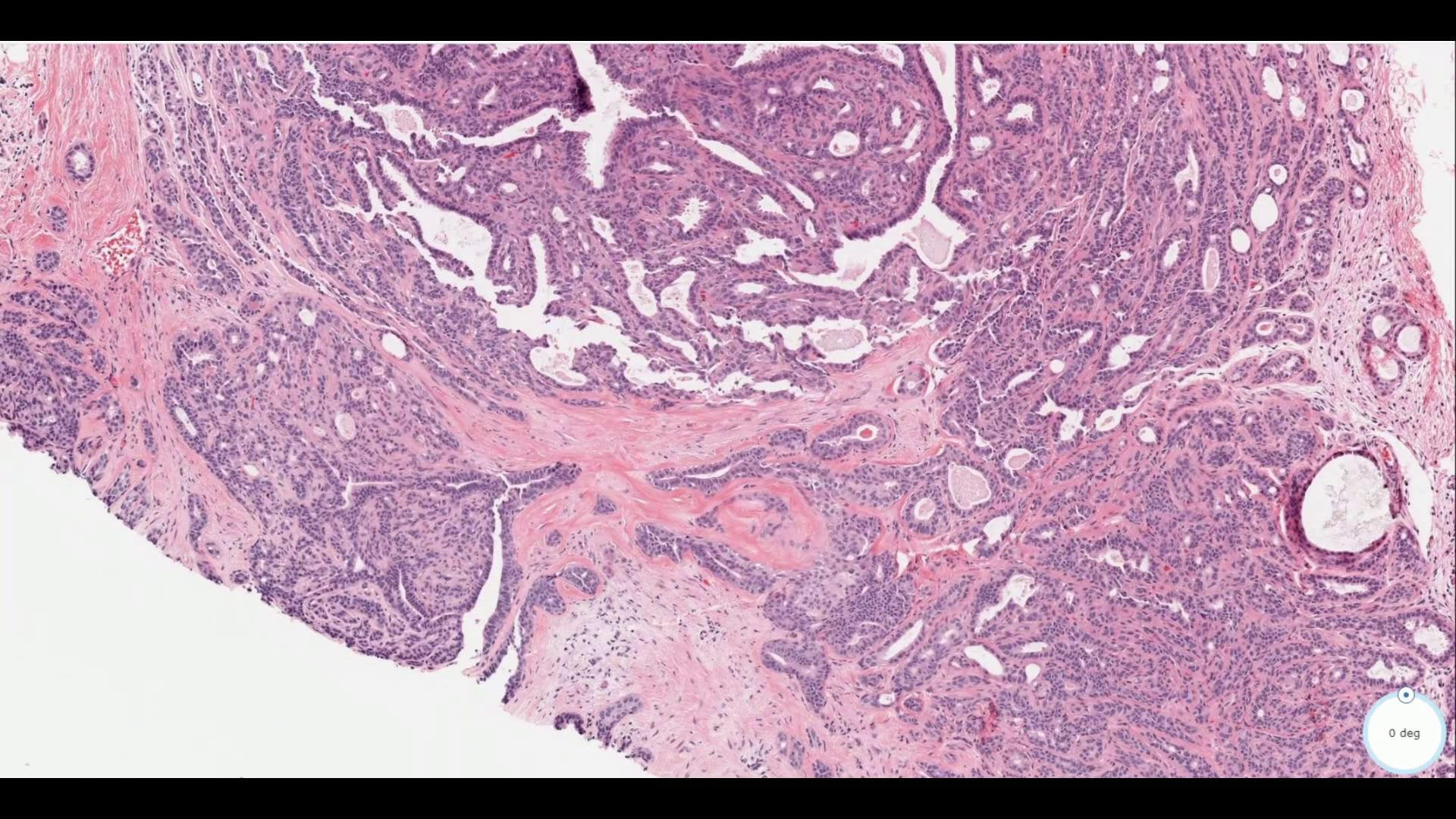} \\
User & How would you describe the stroma visible in the image?\\
 \midrule
\model & The stroma in the image appears to be quite cellular, indicating a high density of cells. This is a significant observation as it can be a sign of certain pathological conditions.\\
\midrule
LLaVA-Med & The stroma in the image appears to be moderately cellular. The stroma is the supportive tissue surrounding the tumor cells, and its cellularity refers to the density of the cells within the stroma. In this case, the stroma is described as moderately cellular, which means it has a moderate amount of cells compared to a highly cellular or poorly cellular stroma.
\\
\midrule
LLaVA & The stroma in the image is described as being "pink" and "red," which suggests that it is a mixture of pink and red tissue.
 \\
 \midrule
Answer &  The stroma visible in the image is described as sclerotic.
 \\
\bottomrule
\end{tabular}
}
\vspace{1mm}
\captionof{table}{\color{red} Failure case \color{black}: Example comparison of histopathology visual chat and reasoning capabilities. }
\vspace{-5mm}
\label{tab:fails_visual_chat_example3_appendix}  
  \end{minipage}
\end{table*}

\setcounter{table}{11}
\begin{table*}[h!]
  \begin{minipage}{0.99\textwidth}
\centering  
\vspace{-4mm}
\scalebox{0.88}{
\begin{tabular}{l p{12.5cm} }
\toprule
 \multicolumn{2}{l}{\bf Visual input example, Histopathology image:}  \\
\midrule
&  \includegraphics[height=3.5cm]{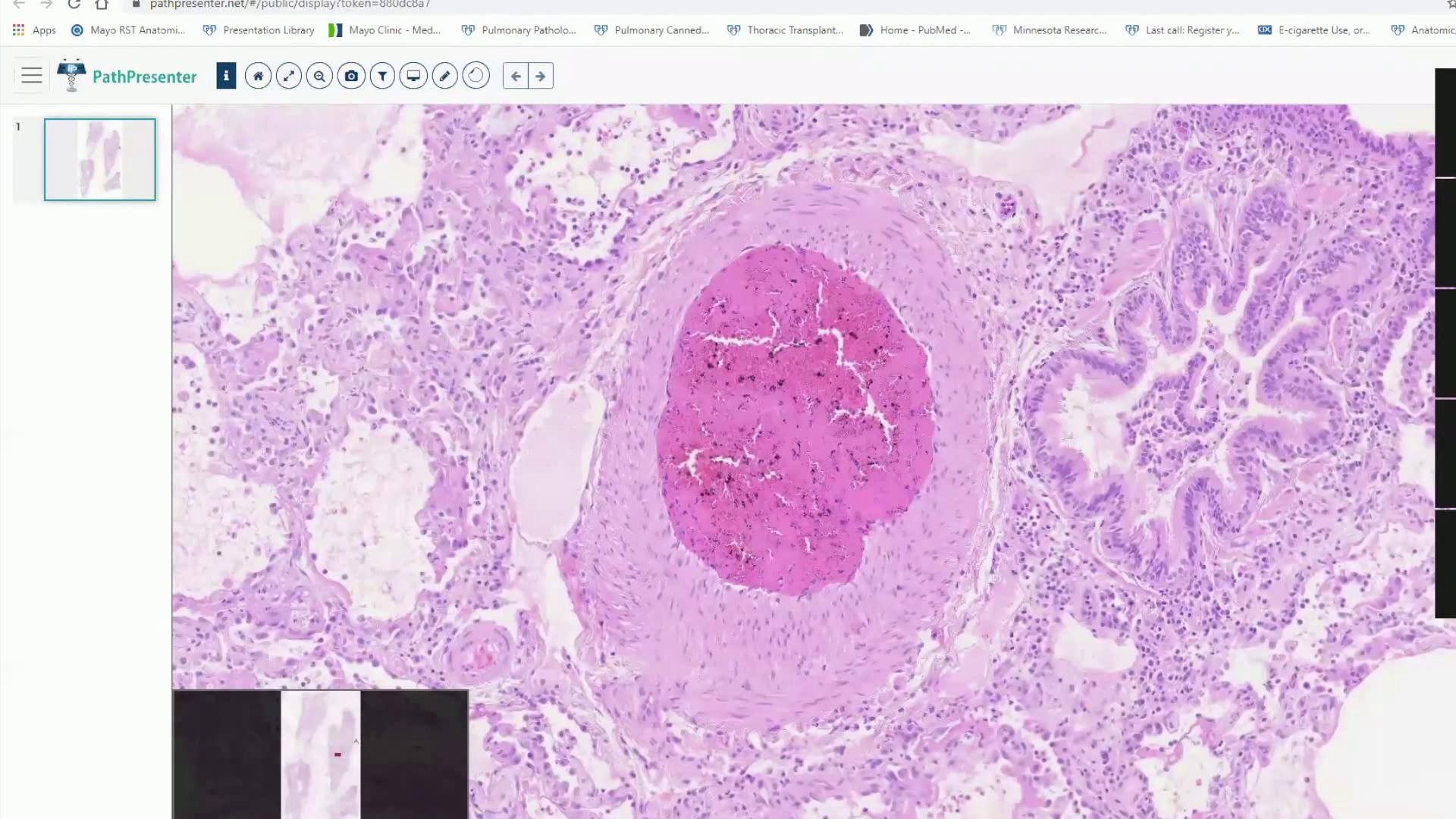} \\
User & How would you characterize the structures lining the alveolar spaces in the image?
\\
 \midrule
\model & The structures lining the alveolar spaces in the image are quite distinctive. They are not typical of what you would expect to see in a healthy lung tissue. Instead, they resemble a certain type of growth or formation, which is not commonly seen in normal lung histology.
\\
\midrule
LLaVA-Med & The structures lining the alveolar spaces in the image are characterized as delicate and flattened.
\\
\midrule
LLaVA &  The structures lining the alveolar spaces in the image are described as "pink" and "purple." These colors suggest that the structures are likely composed of blood vessels, which are responsible for delivering oxygen to the body's cells. The presence of these blood vessels in the image indicates that the tissue is actively involved in the process of respiration and oxygen exchange.
 \\
 \midrule
Answer &  The structures lining the alveolar spaces in the image are isoenophilic hyaline membranes. They are classic features of a diffuse alveolar damage process.
 \\
\bottomrule
\end{tabular}
}
\vspace{1mm}
\captionof{table}{\color{red} Failure case \color{black}: Example comparison of histopathology visual chat and reasoning capabilities. }
\vspace{-5mm}
\label{tab:fails_visual_chat_example4_appendix}  
  \end{minipage}
\end{table*}

\subsection{Omission of GPT-4V in evaluation}

Due to the nature of the test datasets being crawled from videos (e.g \quiltvqa) and articles (e.g. PMC-VQA-Subset) the images most often contain confounding artifacts which can often be indicative of diagnosis, or image features in text, we find that GPT-4V, unlike LLAVA models, are better capable at extracting tiny textual confounding artifacts that describe the image, when such telling artifacts are within the image and would otherwise fail when the artifacts are removed, See Table \ref{tab:gpt4v_appendix}, based on this observation we exclude GPT-4V from our evaluations and leave its evaluation to future work.

\setcounter{table}{12}
\begin{table*}
  \begin{minipage}{0.99\textwidth}
\centering  
\vspace{-4mm}
\scalebox{0.88}{
\begin{tabular}{l p{12.5cm} }
\toprule
 \multicolumn{2}{l}{\bf GPT-4V example, Histopathology image:}  \\
\midrule
&  \includegraphics[height=5.5cm]{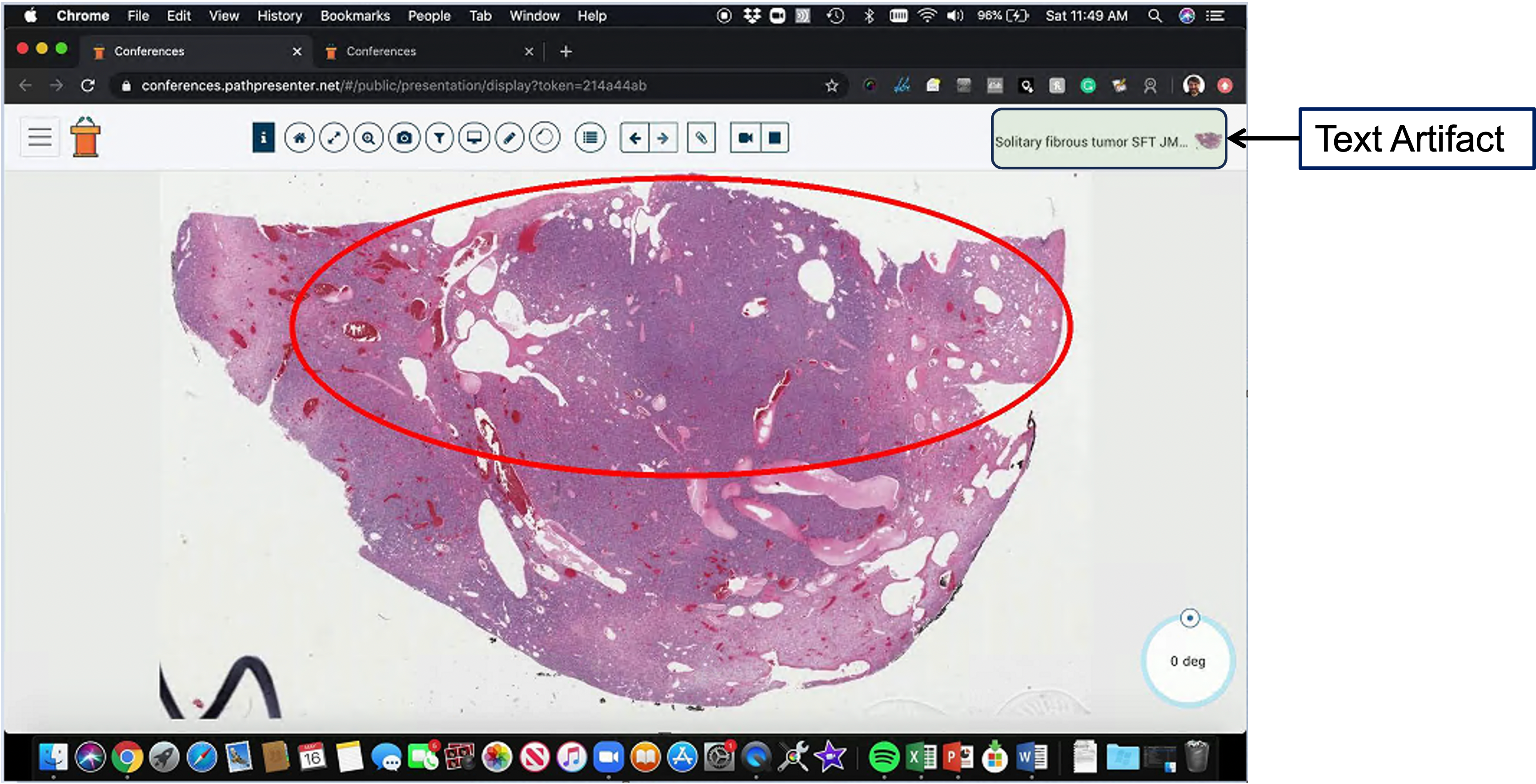} \\
&   \includegraphics[height=3.5cm]{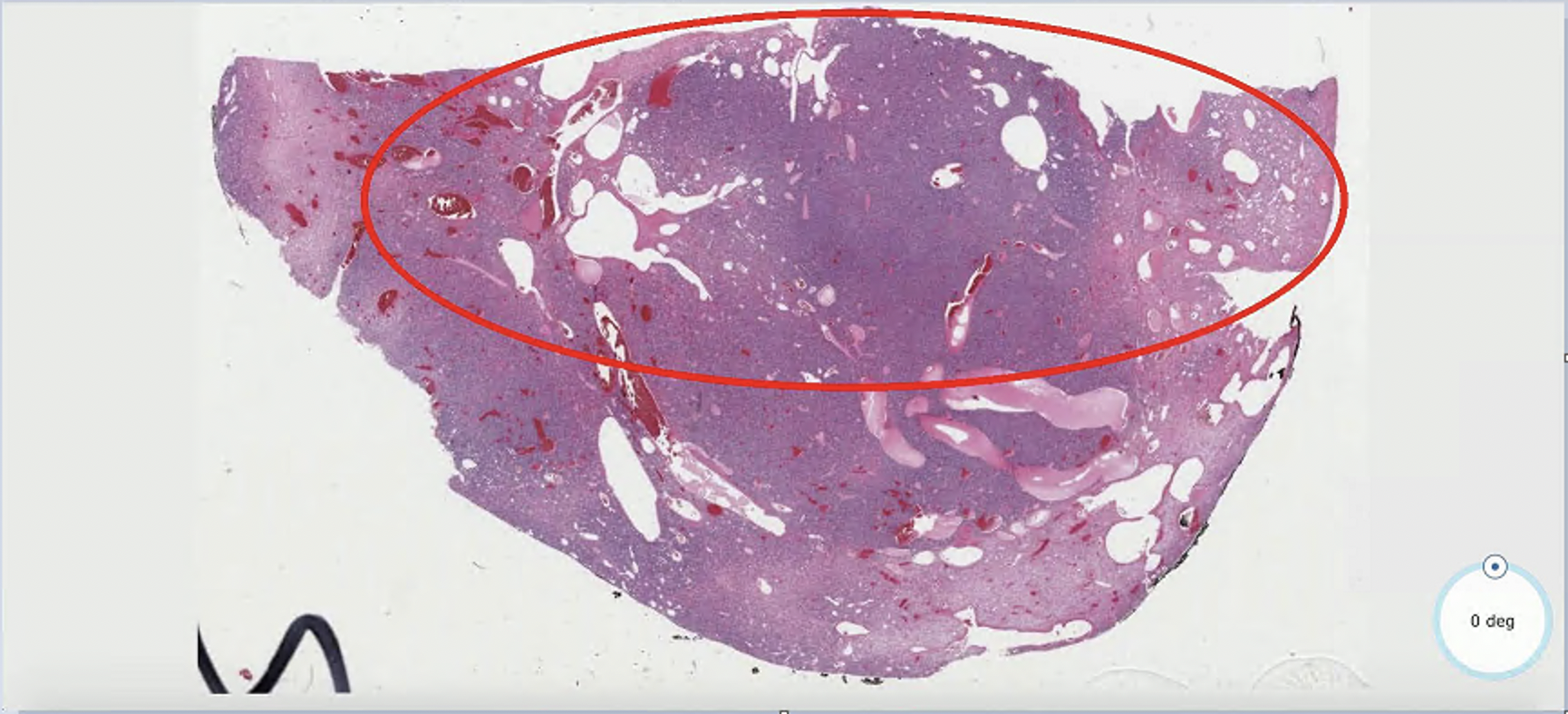} \\
User & What is the unique vascular pattern seen in the image? Tip: Focus on image areas highlighted by the red circle(s) \\
 \midrule
\model & The image reveals a unique vascular pattern that is quite distinctive. The red circles, which are highlighted, seem to be pointing towards a specific area of interest. This area is characterized by a dense network of blood vessels, which is a common feature in certain types of tumors. The vessels appear to be irregularly shaped and densely packed, which is not typically seen in normal tissue. This vascular pattern could be indicative of a highly vascularized tumor, which is often associated with aggressive growth and poor prognosis.
\\
\midrule
GPT-4V w/o Cropping & The image you've provided appears to be a histological slide of tissue, most likely viewed under a microscope and stained for better visualization of the structures. The red circle highlights an area of interest within the tissue. The unique vascular pattern within the red circle seems to be characteristic of a "staghorn" or "hemangiopericytoma-like" vascular pattern. This pattern is often associated with solitary fibrous tumors (SFTs), which are rare tumors that can occur in various parts of the body. The vessels in this pattern tend to be elongated and branched, resembling the antlers of a stag, hence the name "staghorn."
Solitary fibrous tumors are typically composed of spindle-shaped cells and have this distinctive vascular pattern. They can be benign or malignant and may require further clinical correlation and immunohistochemical staining to confirm the diagnosis and plan appropriate treatment. Please note that the interpretation of histological images should be performed by a trained pathologist, and this assessment is based on the visual information provided in the image.
\\
\midrule
GPT-4V w/ Cropping  &  The image appears to be a microscopic slide of a tissue sample, which has been stained for better visualization of the structures within the tissue. The red circle drawn on the image seems to highlight a particular area of interest, possibly for educational or diagnostic purposes. The tissue exhibits various shades of purple and pink, indicating different types of cells and extracellular components. The presence of white spaces could represent adipose (fat) tissue, voids where cells are not present, or artifacts from the slide preparation process. The label "0 deg" in the corner suggests that the image may be part of a series taken at different rotations or angles. Without more context, it's difficult to determine the exact type of tissue or the purpose of the analysis.
 \\
 \midrule
Answer &  The unique vascular pattern seen in the image is characteristic of solitary fibrous tumors, previously known as hemangiopericytomas. This pattern includes dilated, branching, tortuous vessels.
 \\
\bottomrule
\end{tabular}
}
\vspace{1mm}
\captionof{table}{Example evaluation GPT-4V: The textual confounding artifact highlighted with green box: {\it solitary fibrous tumor SFT JM..} is picked up by GPT-4 and forms the basis of its description w/o cropping, however, when the text is cropped out of the image, it is clear that GPT-4V lacks histology image understanding as it simply describes the colors and other non-confounding artifacts missing the histology features within the image e.g the vessel structures.}
\label{tab:gpt4v_appendix}  
  \end{minipage}
\end{table*}

\setcounter{figure}{4}
\begin{figure*}[h!]
\begin{center}
\includegraphics[width=1\linewidth]{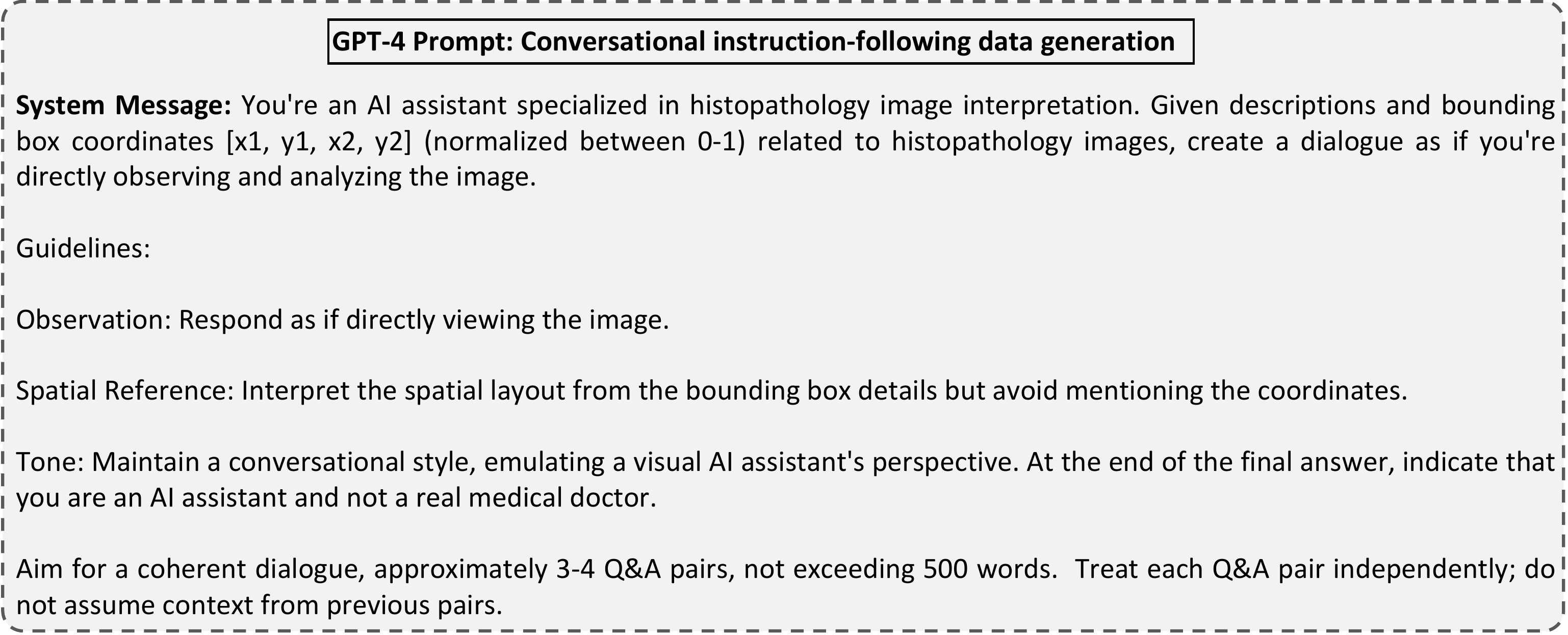}
\end{center}
   \caption{The GPT-4 prompt used to generate conversational instruction-following data.}
\label{fig:prompt_conversational}
\end{figure*}

\setcounter{figure}{5}
\begin{figure*}[h!]
\begin{center}
\includegraphics[width=1\linewidth]{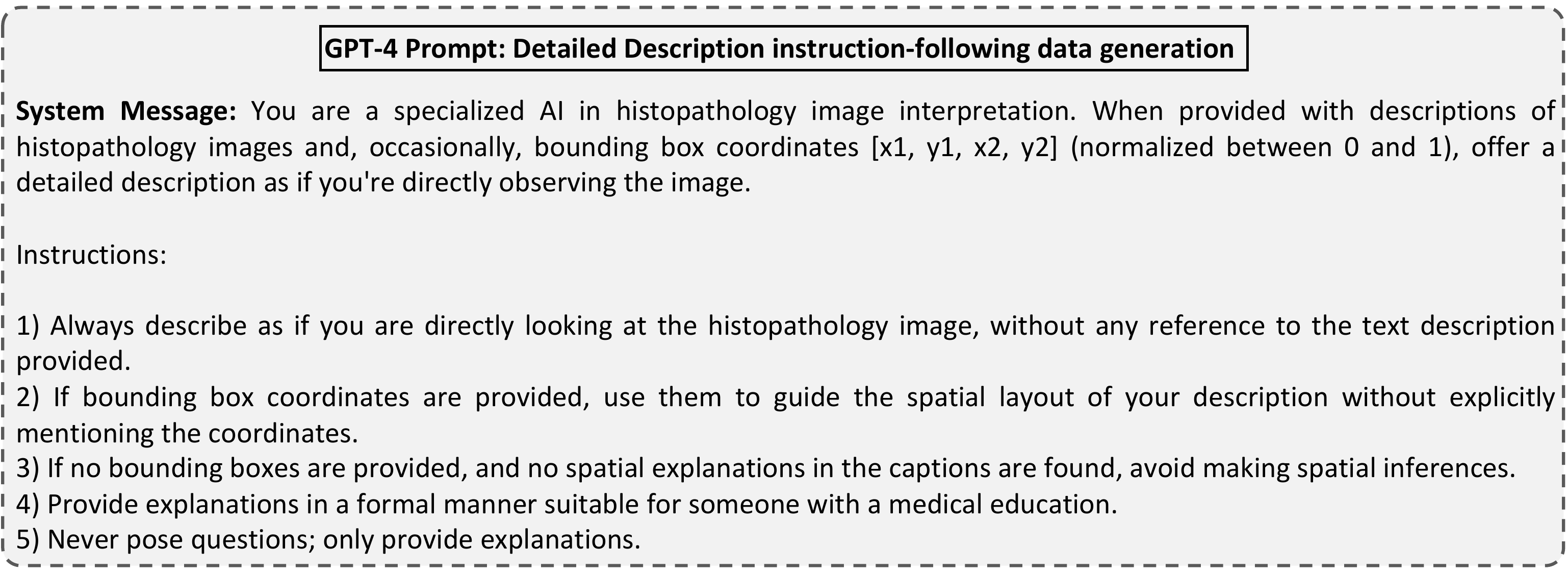}
\end{center}
   \caption{The GPT-4 prompt used to generate detailed description instruction-following data.}
\label{fig:prompt_description}
\end{figure*}

\setcounter{figure}{6}
\begin{figure*}[h!]
\begin{center}
\includegraphics[width=1\linewidth]{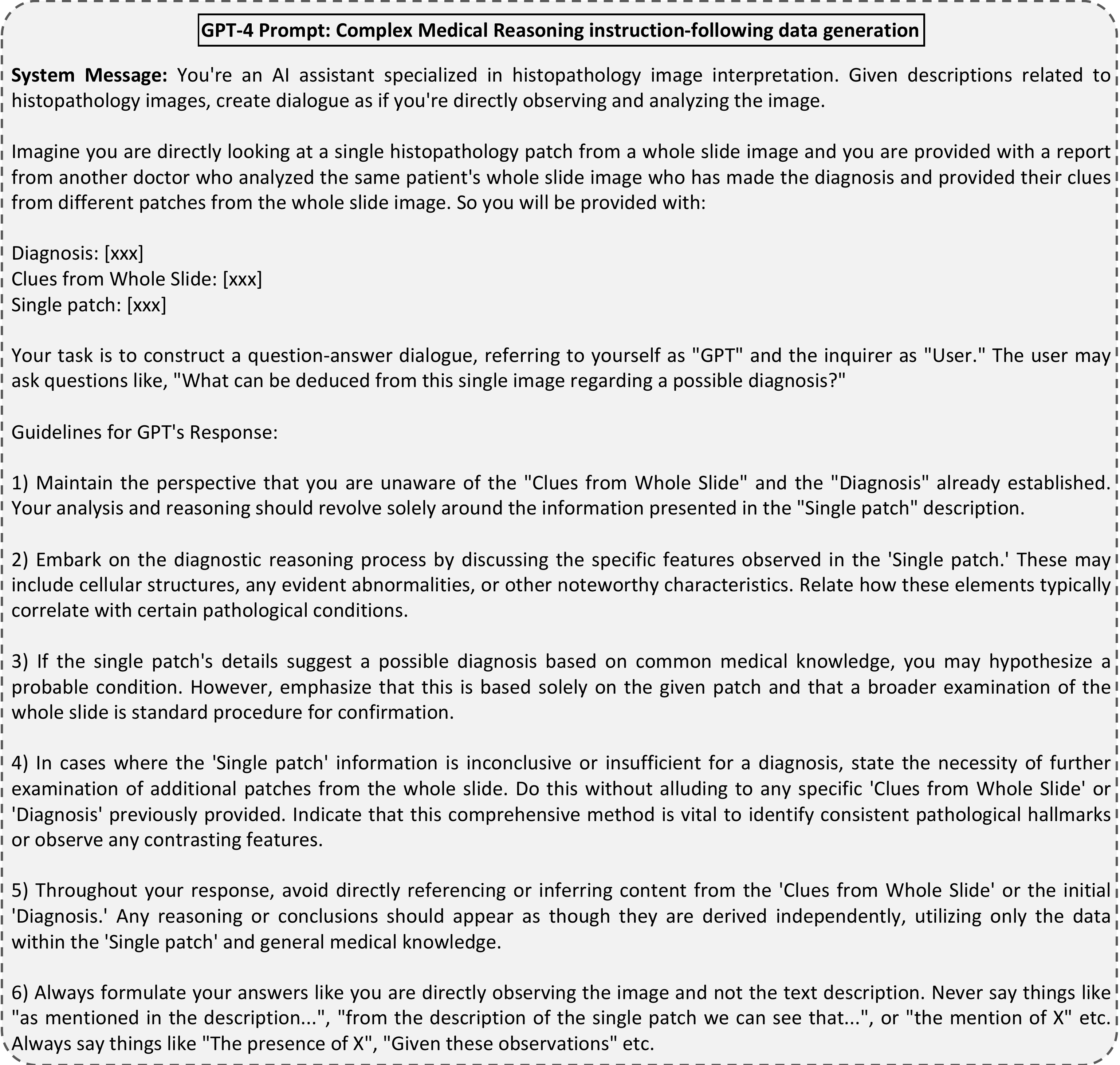}
\end{center}
   \caption{The GPT-4 prompt used to generate complex medical reasoning instruction-following data.}
\label{fig:prompt_complex}
\end{figure*}

\setcounter{figure}{7}
\begin{figure*}[h!]
\begin{center}
\includegraphics[width=1\linewidth]{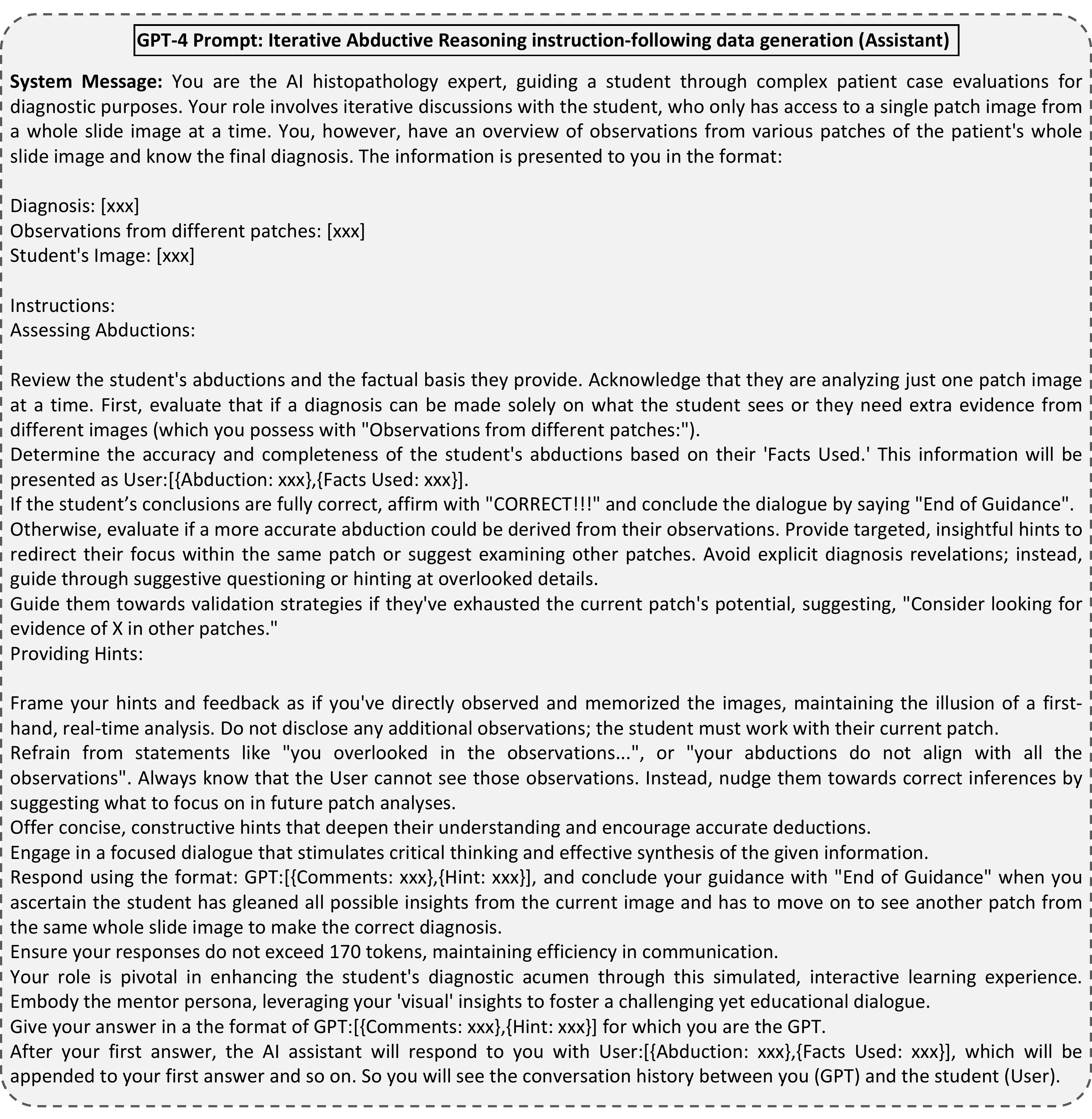}
\end{center}
   \caption{The GPT-4 prompt used to generate iterative abductive reasoning (assistant) instruction-following data.}
\label{fig:prompt_iterative_assistant}
\end{figure*}

\setcounter{figure}{8}
\begin{figure*}[ht]
\begin{center}
\includegraphics[width=1\linewidth, height=0.95\textheight]{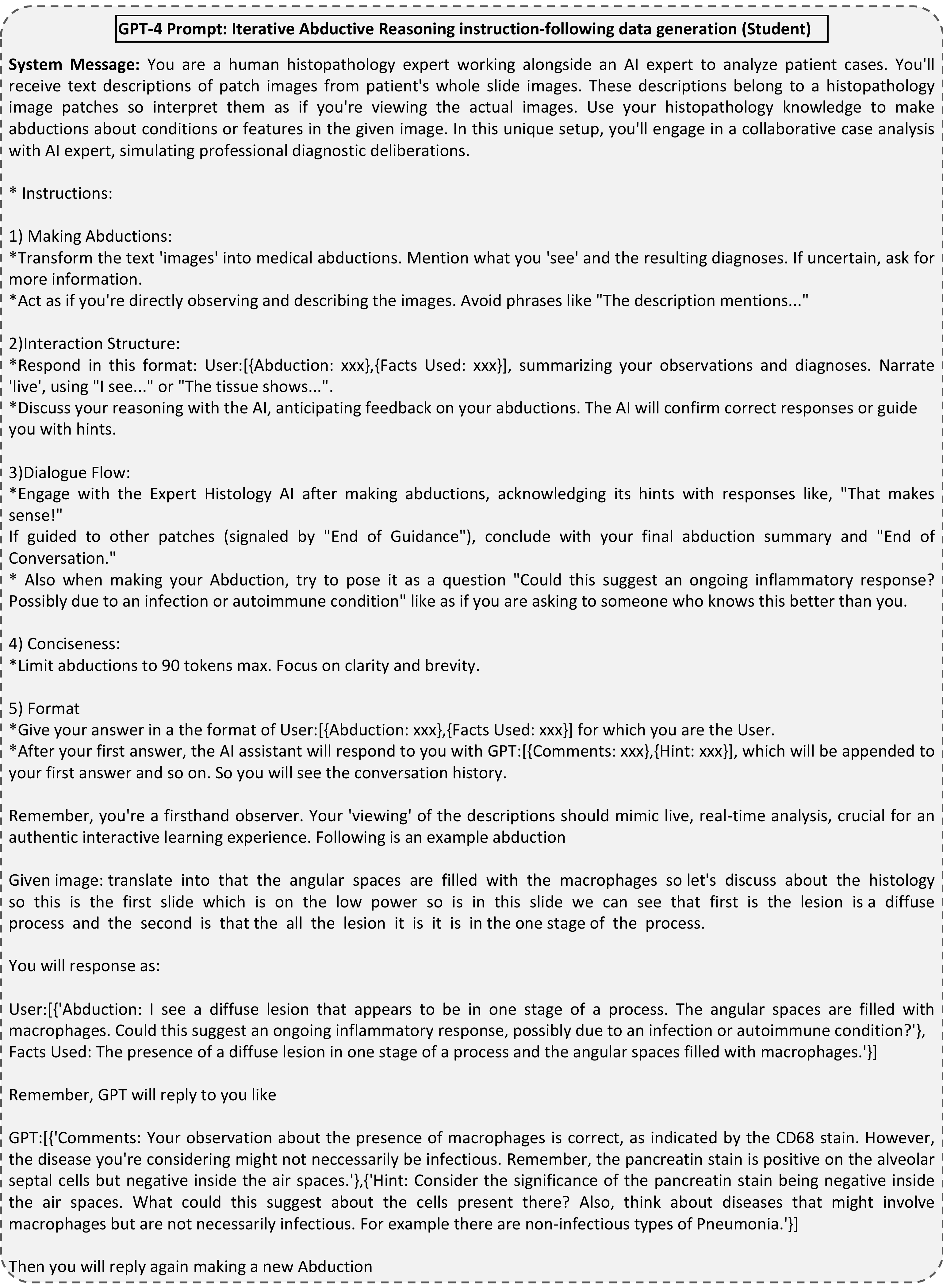}
\end{center}
   \caption{The GPT-4 prompt used to generate iterative abductive reasoning (student) instruction-following data.}
\label{fig:prompt_iterative_student}
\end{figure*}

\setcounter{figure}{9}
\begin{figure*}[ht]
\begin{center}
\includegraphics[width=1\linewidth]{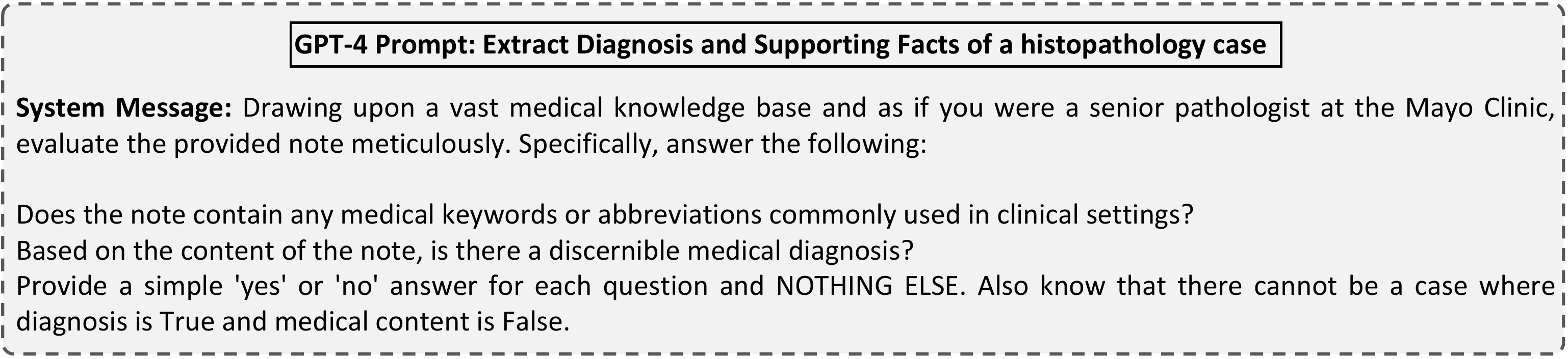}
\end{center}
   \caption{The GPT-4 prompt used to extract the diagnosis and supporting facts of a histopathology case.}
\label{fig:diag_prompt}
\end{figure*}

\setcounter{figure}{10}
\begin{figure*}[ht]
\begin{center}
\includegraphics[width=1\linewidth]{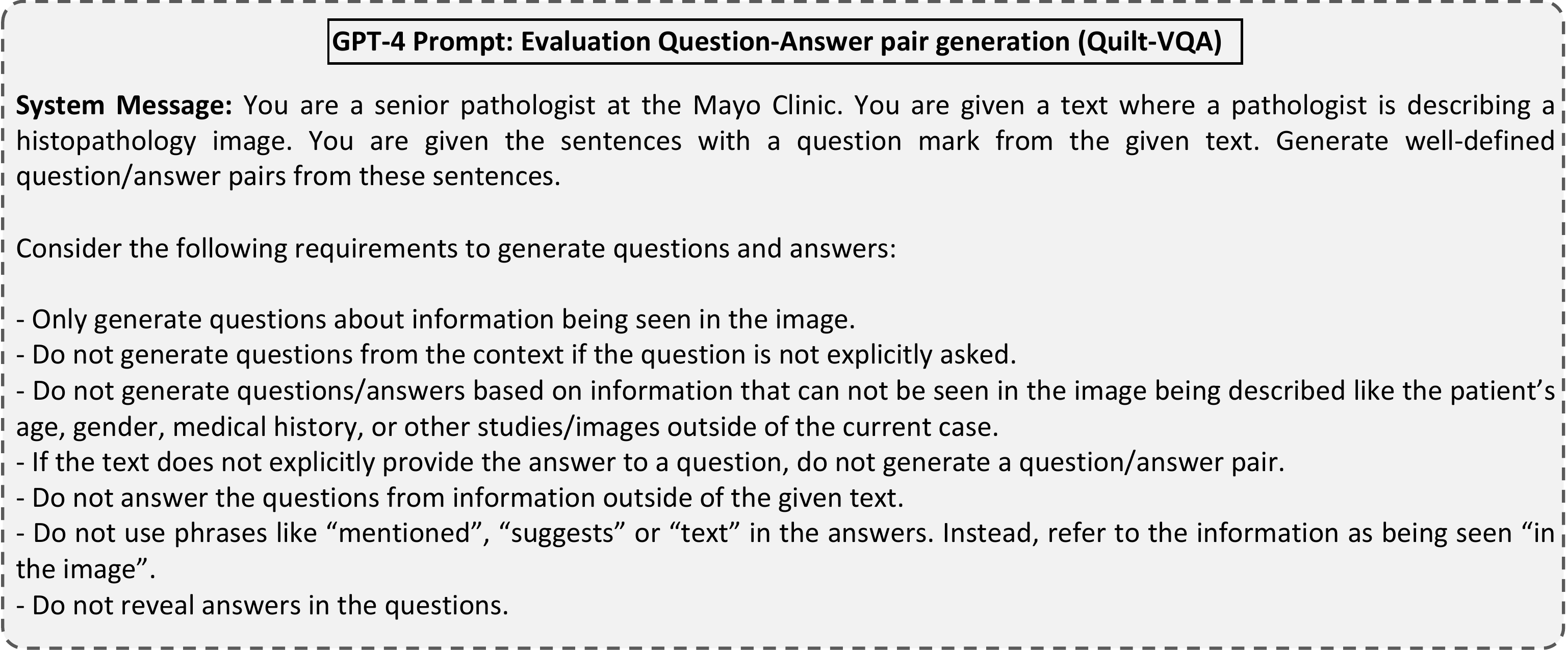}
\end{center}
   \caption{The GPT-4 prompt used to generate question-answer pairs for evaluation set (\quiltvqa).}
\label{fig:eval_prompt}
\end{figure*}

\setcounter{figure}{11}
\begin{figure*}[ht]
\begin{center}
\includegraphics[width=1\linewidth]{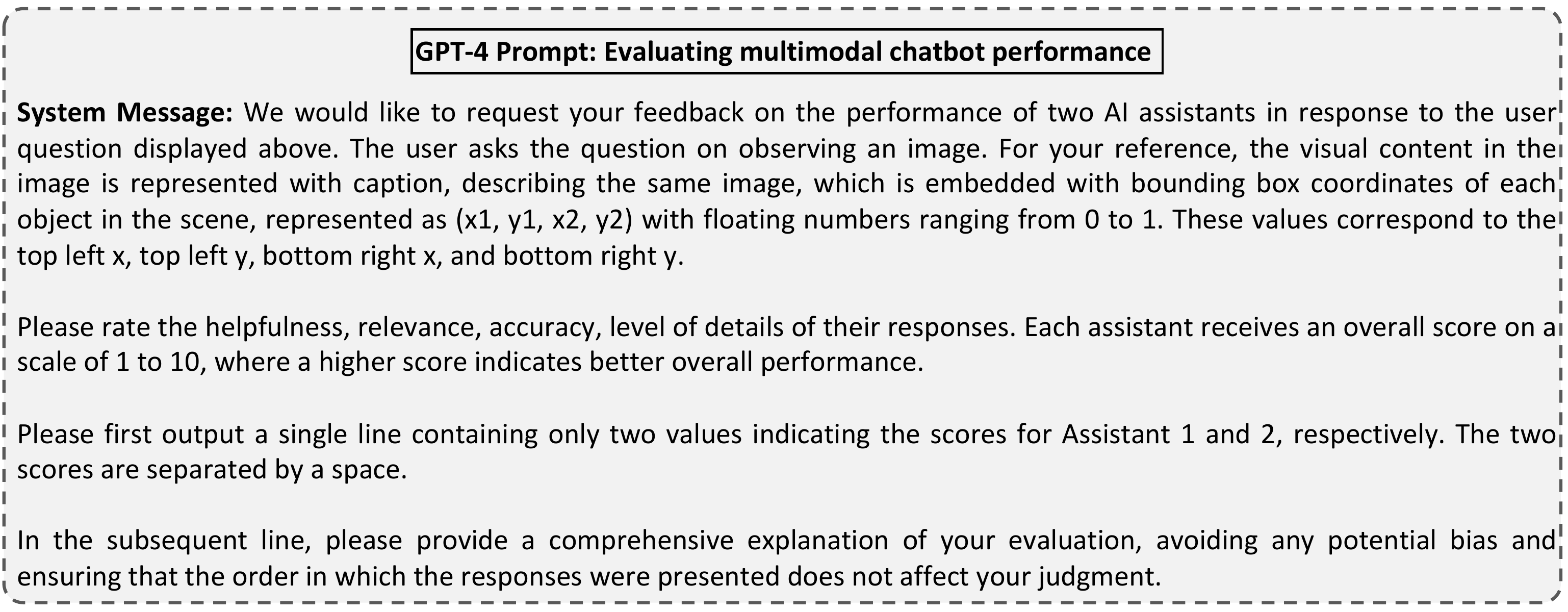}
\end{center}
   \caption{The GPT-4 prompt used to evaluate the multi-modal model performance.}
\label{fig:eval_gpt_score}
\end{figure*}


\setcounter{figure}{12}
\begin{figure*}[ht!]
\begin{center}
\includegraphics[width=1\linewidth]{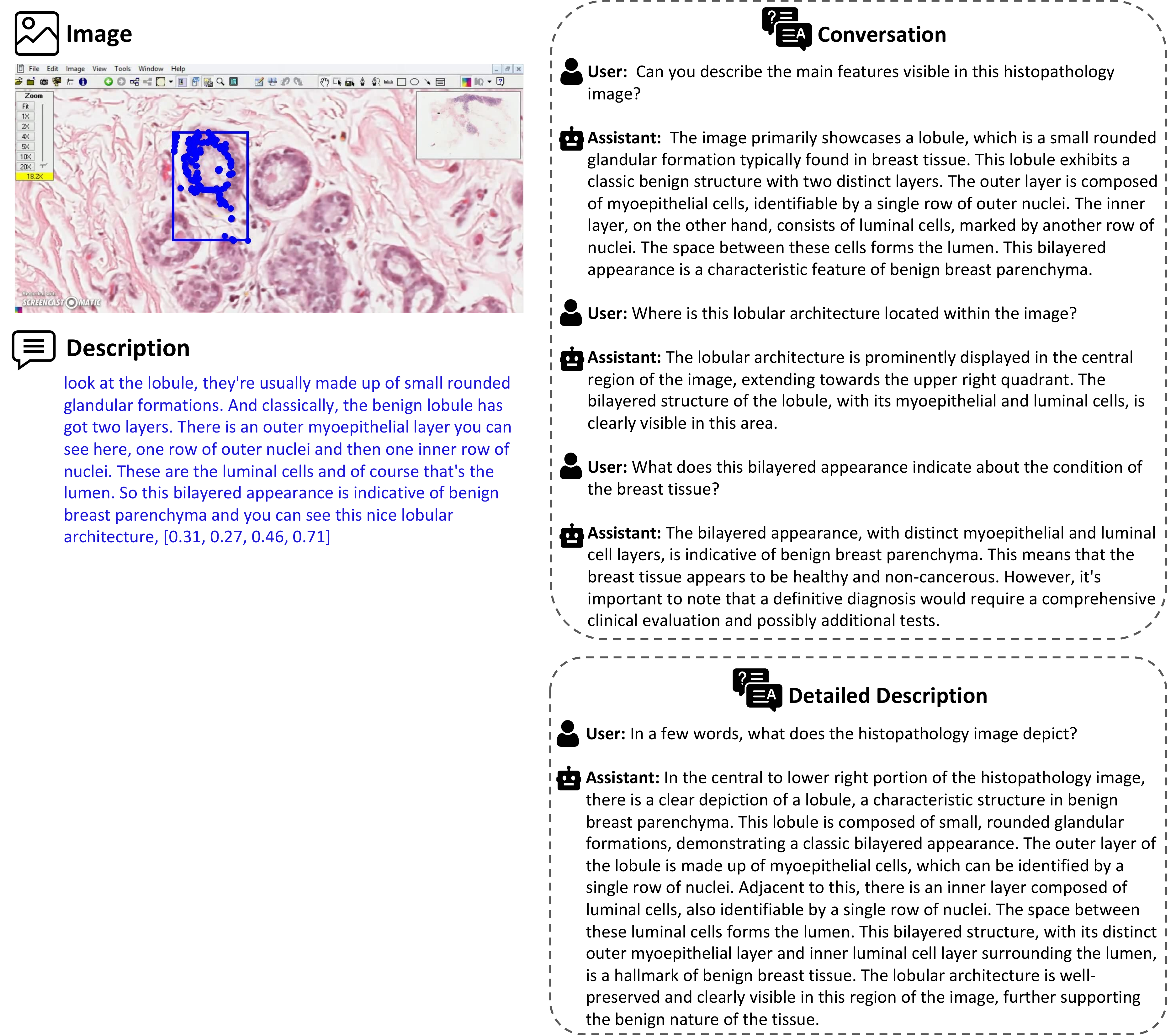}
\end{center}
   \caption{A full example of Conversation and Detailed Description types of Q/A in \quiltinstruct.}
\label{fig:qa_1_2}
\end{figure*}

\setcounter{figure}{13}
\begin{figure*}[ht!]
\begin{center}
\includegraphics[width=1\linewidth]{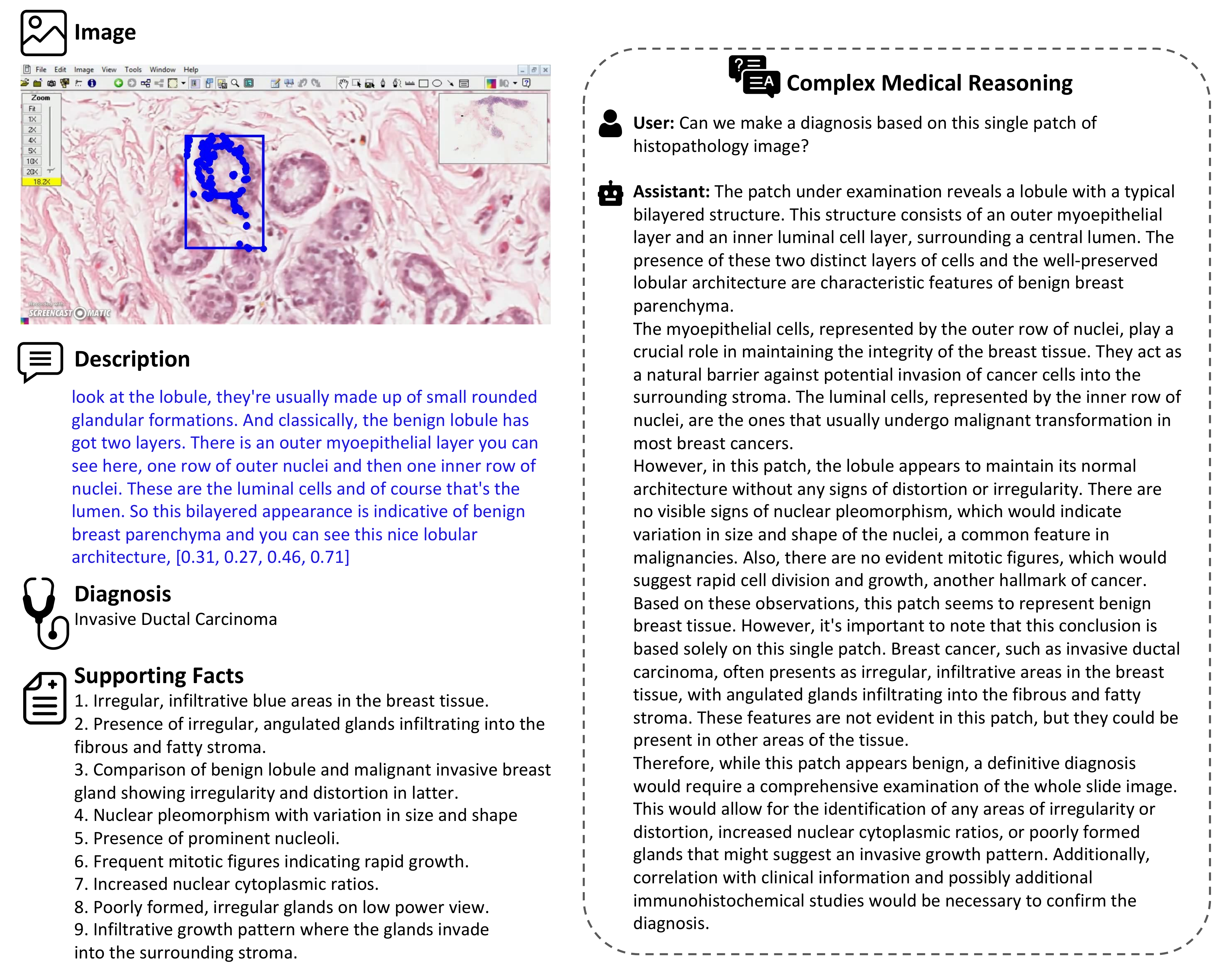}
\end{center}
   \caption{A full example of Complex Medical Reasoning type of Q/A in \quiltinstruct.}
\label{fig:qa_3}
\end{figure*}

\setcounter{figure}{14}
\begin{figure*}[ht!]
\begin{center}
\includegraphics[width=1\linewidth]{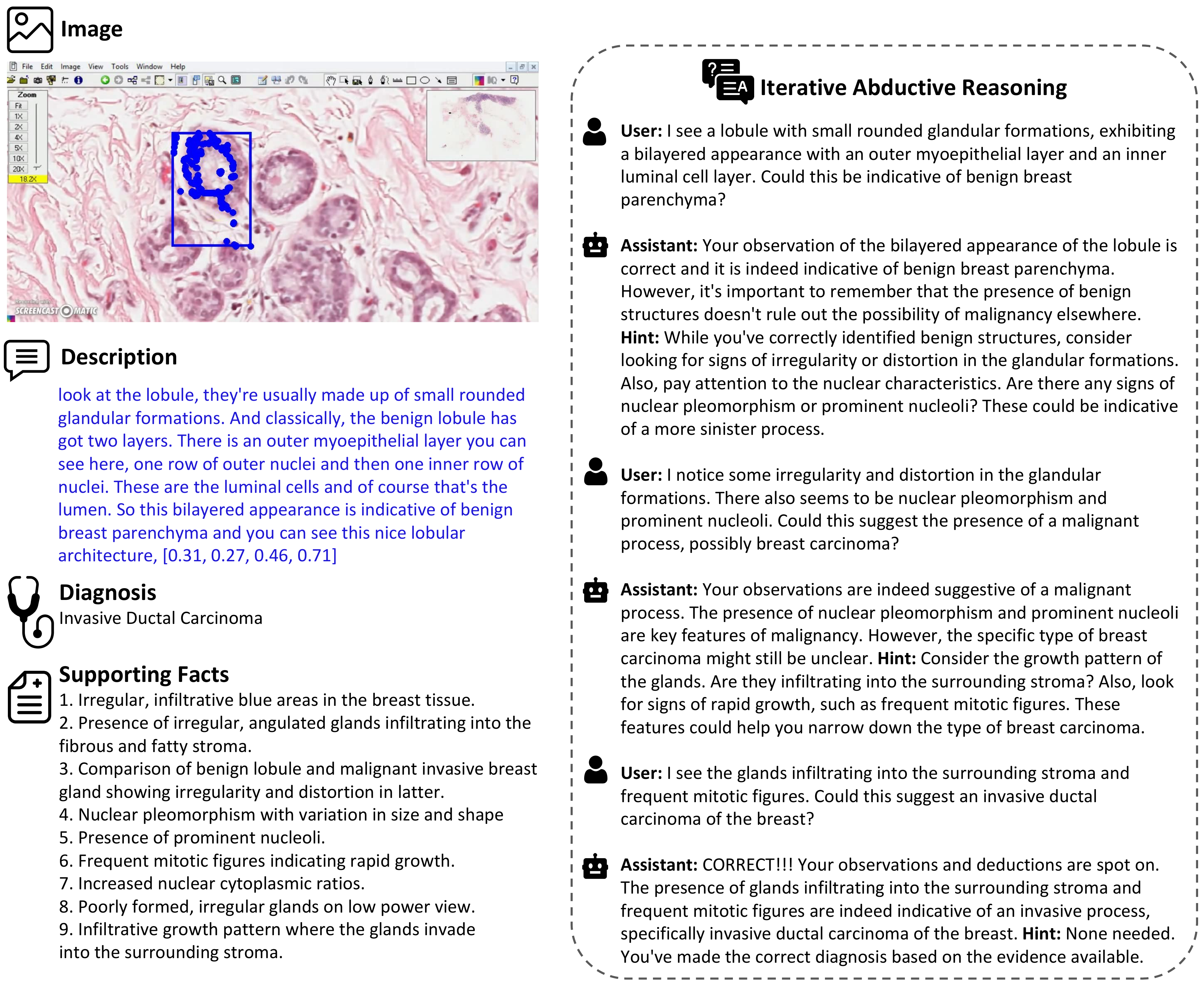}
\end{center}
   \caption{A full example of Iterative Abductive Reasoning type of Q/A in \quiltinstruct.}
\label{fig:qa_4}
\end{figure*}

\setcounter{figure}{15}
\begin{figure}[ht!]
\begin{center}
\includegraphics[width=1\linewidth]{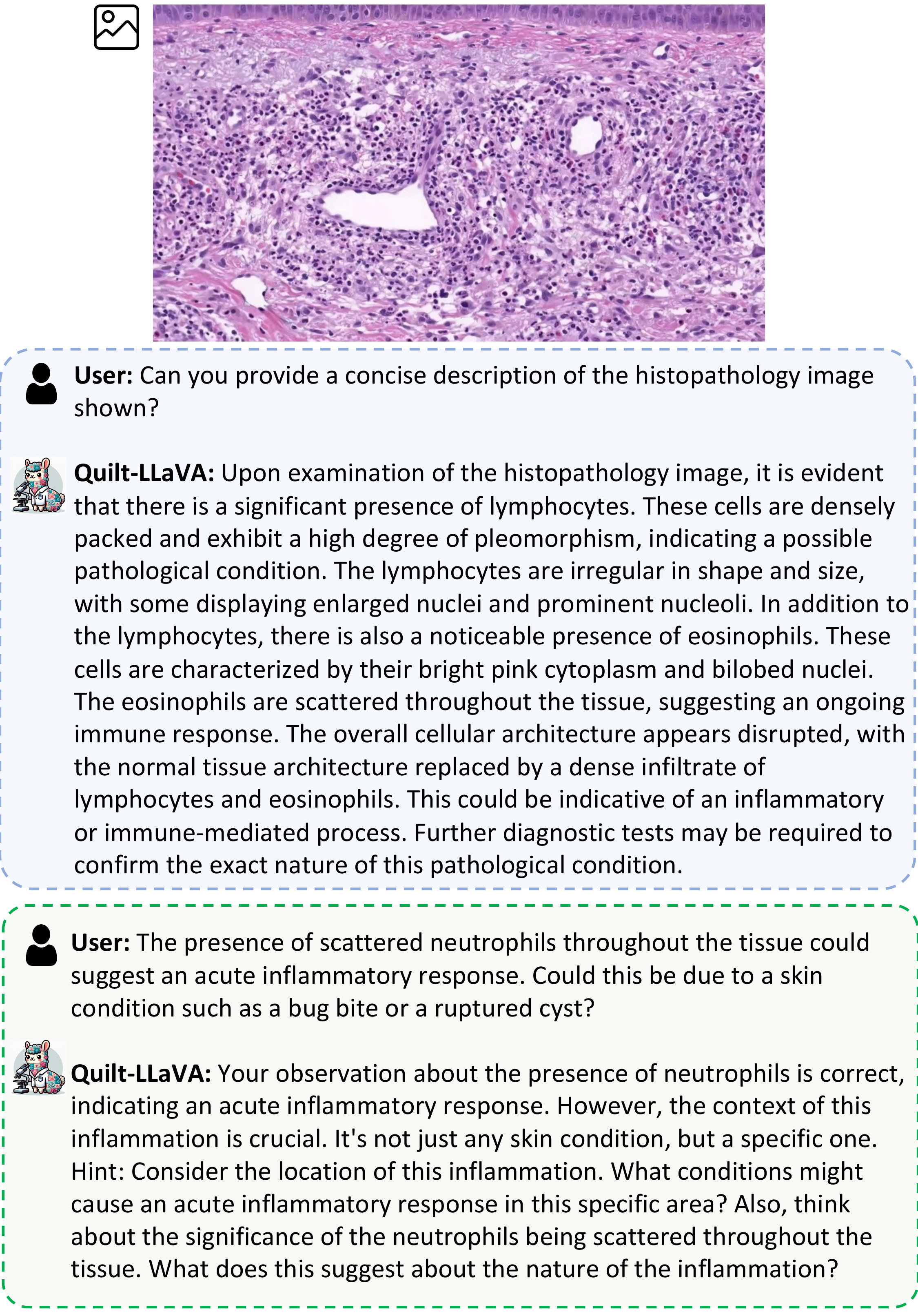}
\end{center}
   \caption{An example illustrating where the \model is used as an educational companion, providing hints to human users to help train them as pathologists.}
\label{fig:educational_reasoning}
\end{figure}

\setcounter{figure}{17}
\begin{figure}[ht!]
\begin{center}
\includegraphics[width=1\linewidth,height=0.2\textheight]{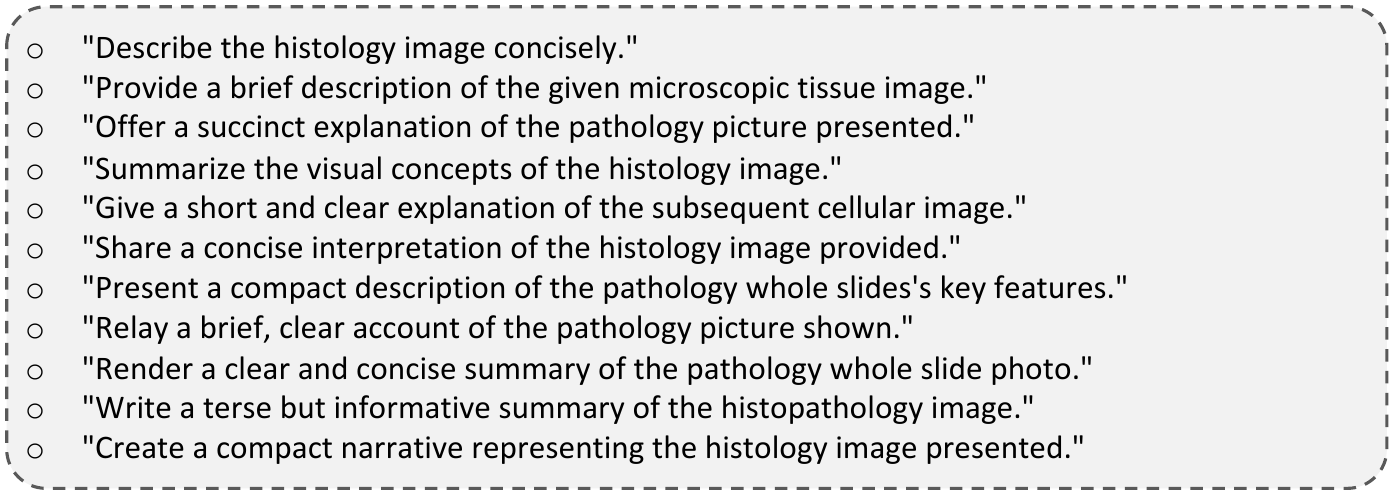}
\end{center}
   \caption{A list of instructions to variably describe the visual information present in the images used in the histopathology domain alignment. The list is taken from \cite{li2023llava} and adapted for the histopathology content.}
\label{fig:question_list_stage1}
\end{figure}

\setcounter{figure}{18}
\begin{figure}[ht!]
\begin{center}
\includegraphics[width=1\linewidth,height=0.2\textheight]{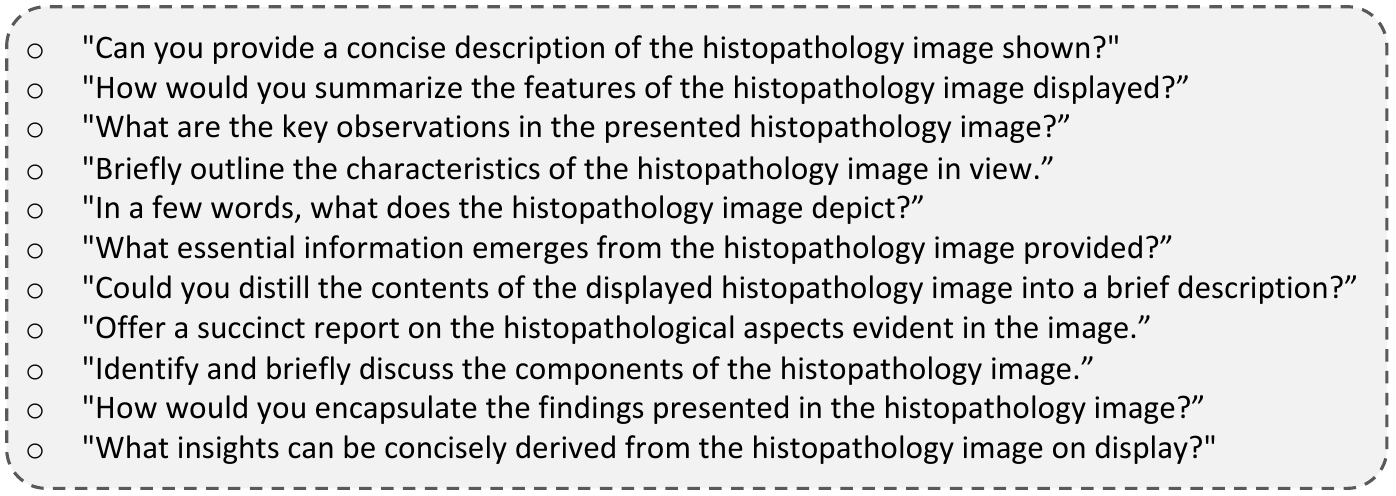}
\end{center}
   \caption{A list of instructions to variably describe the visual information present in the images in detail used to generate detailed description instruction-following QA pairs.}
\label{fig:detailed_description_list}
\end{figure}

\setcounter{figure}{19}
\begin{figure*}[h!]
\begin{center}
\includegraphics[width=1\linewidth]{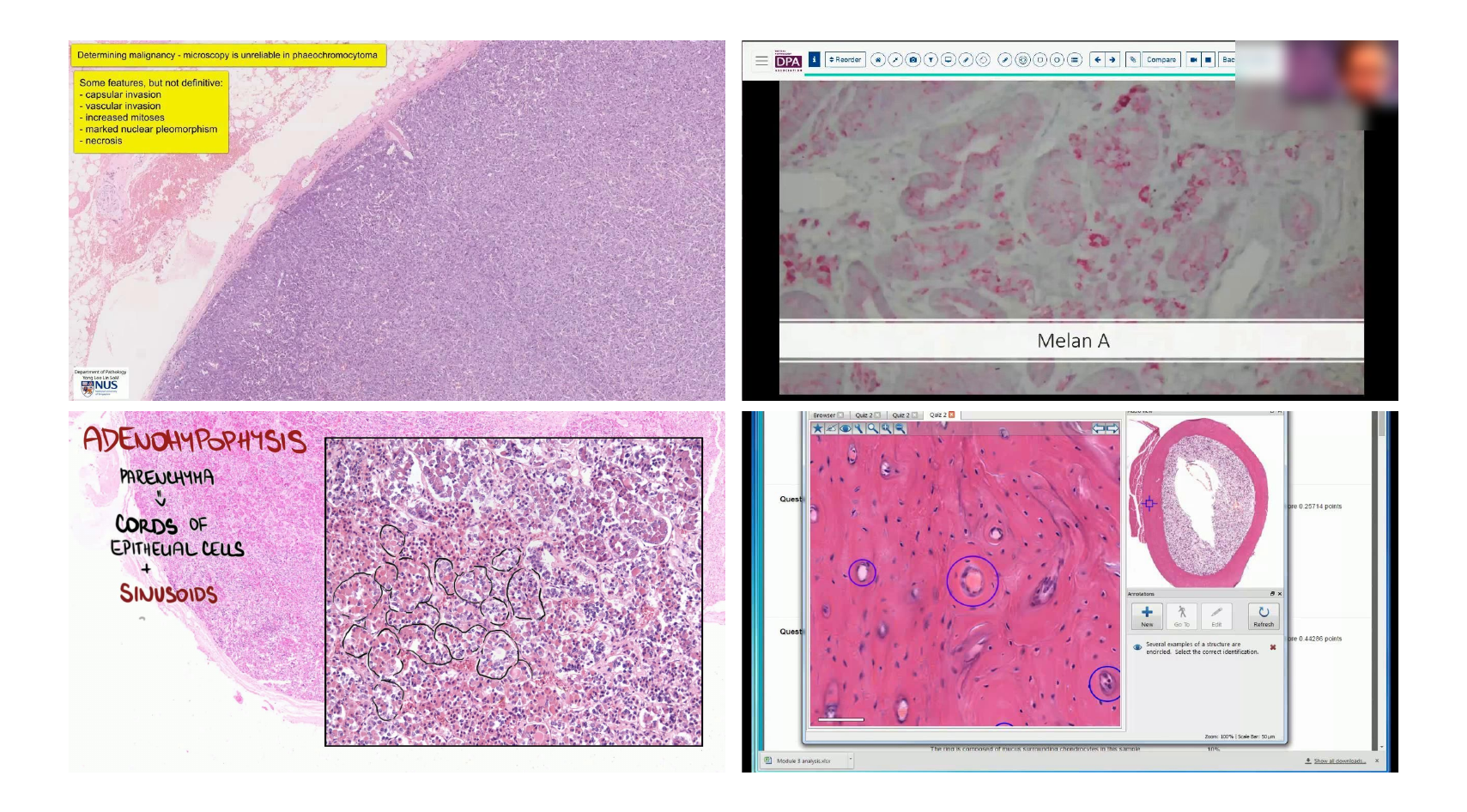}
\end{center}
   \caption{Some examples of noisy images include those where the median frame might contain artifacts such as text, drawings, or the narrator's face (seen in the top right figure). Although we completely mask the narrator's face in our process, for brevity, we have included an example here with a blurred face.)}
\label{fig:image_artifacts}
\end{figure*}

\setcounter{figure}{20}
\begin{figure*}[h!]
\begin{center}
\includegraphics[width=1\linewidth]{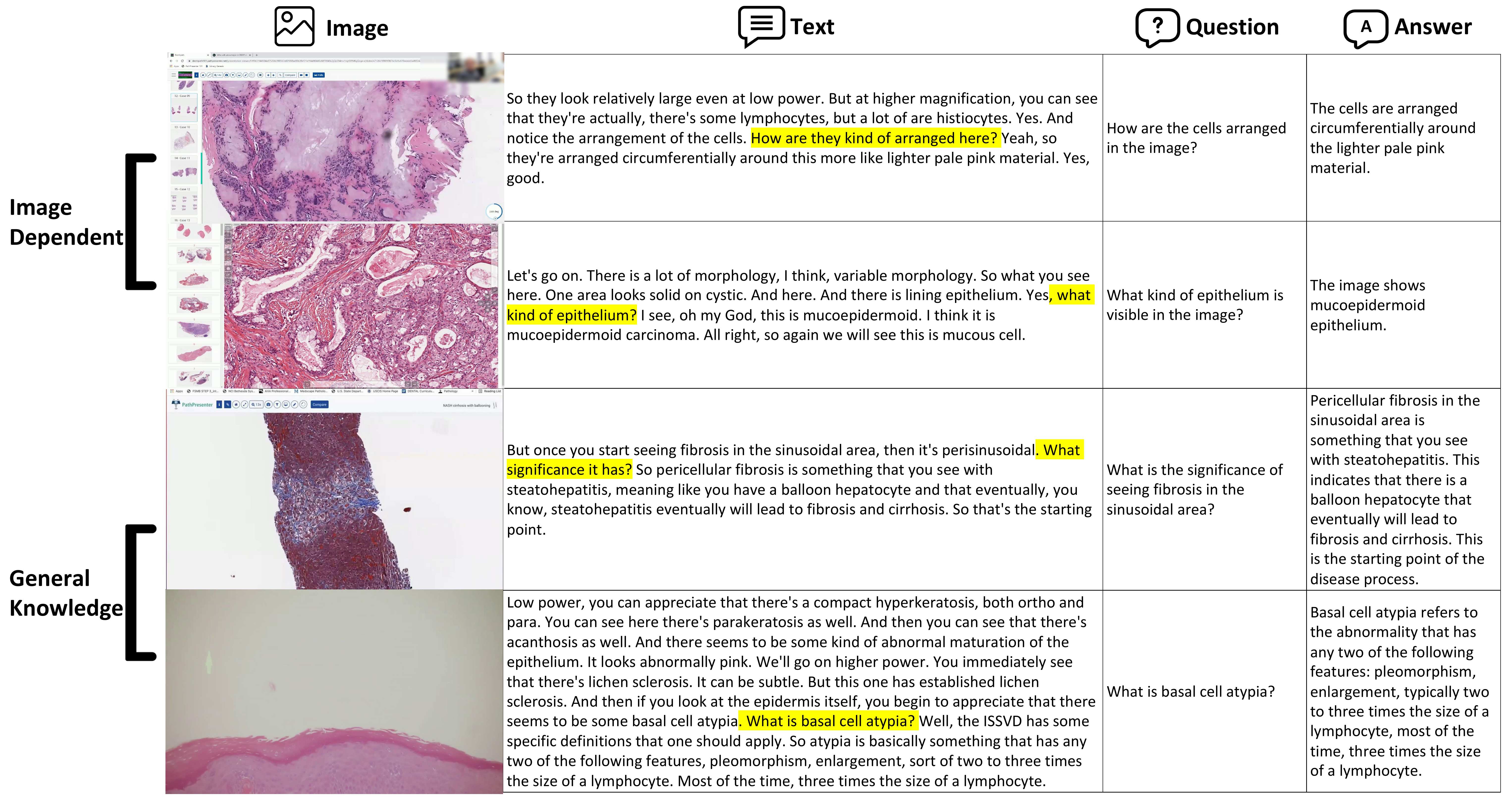}
\end{center}
   \caption{Some examples of \quiltvqa, our evaluation dataset. Top two rows show image-dependent Q/A pairs and bottom two rows show general-knowledge Q/A pairs. The original question posed by the narrator of the video is highlighted in yellow.}
\label{fig:quilt_vqa_samples}
\end{figure*}


\end{document}